\documentclass[preprint]{elsarticle}

\usepackage{amsmath}
\usepackage{amssymb}
\usepackage{latexsym}

\usepackage{subfig}
\usepackage{graphicx}
\usepackage{epstopdf}

\def\PSML{Parallel Stroked Multi Line}
\def\psml{PSML}
\def\Holder{H{\"o}lder}

\newdefinition{definition}{Definition}
\newtheorem{theorem}{Theorem}
\newtheorem{algorithm}{Algorithm}
\newdefinition{corollary}{Corollary}
\newtheorem{lemma}{Lemma}
\newproof{thmproof}{Proof}

\newlength{\columnwidthe}
\begin{document}

\title{\PSML: a model-based method for compressing large fingerprint databases}
\author[fum]{Hamid~Mansouri\corref{cor1}}\ead{mansouri@stu.um.ac.ir}
\author[fum]{Hamid-Reza~Pourreza\corref{cor2}}\ead{hpourreza@um.ac.ir}
\address[fum]{Machine Vision Lab., Computer Engineering Department, Ferdowsi University of Mashhad, Mashhad 91779, Iran}
\cortext[cor1]{Corresponding author, Tel: +98 915 509 8244}
\cortext[cor2]{Principal corresponding author, Tel: +98 51 3880 5025}

\begin{abstract}
With increasing usage of fingerprints as an important biometric data, the need to compress the large fingerprint databases has become essential. The most recommended compression algorithm, even by standards, is JPEG2000. But at high compression rates, this algorithm is ineffective. In this paper, a model is proposed which is based on parallel lines with same orientations, arbitrary widths and same gray level values located on rectangle with constant gray level value as background. We refer to this algorithm as \PSML\ (\psml). By using Adaptive Geometrical Wavelet and employing \psml, a compression algorithm is developed. This compression algorithm can preserve fingerprint structure and minutiae. The exact algorithm of computing the \psml\ model take exponential time. However, we have proposed an alternative approximation algorithm, which reduces the time complexity to $O(n^3)$. The proposed \psml\ algorithm has significant advantage over Wedgelets Transform in PSNR value and visual quality in compressed images. The proposed method, despite the lower PSNR values than JPEG2000 algorithm in common range of compression rates, in all compression rates have nearly equal or greater advantage over JPEG2000 when used by Automatic Fingerprint Identification Systems (AFIS). At high compression rates, according to PSNR values, mean EER rate and visual quality, the encoded images with JPEG2000 can not be identified from each other after compression. But, images encoded by the \psml\ algorithm retained the sufficient information to maintain fingerprint identification performances similar to the ones obtained by raw images without compression. One the
\textit{U.are.U 400} database, the mean EER rate for uncompressed images is 4.54\%, while at 267:1 compression ratio, this value becomes 49.41\% and 6.22\% for JPEG2000 and PSML, respectively. This result shows a significant improvement over the standard JPEG2000 algorithm.
\end{abstract}


\maketitle

\section{Introduction}
Biometric-based human identification and authentication systems rely and several biometric modality including Fingerprint \cite{finger1}, Palm \cite{Palm1}, Face \cite{Face1}, Iris \cite{Iris1}, Gait \cite{Gait1} and human activity \cite{HumanAct1}. Among these modalities, fingerprint is the most widely used one. Traditionally fingerprint data used to be stored on paper. But, with advancements in machine-based fingerprint analysis, the data is now stored in digital format. Automated Fingerprint Identification Systems (AFIS) utilize the digitized data to perform the matching and verification. These systems are now highly reliable and even European Union (EU) recently decided to use fingerprint data in digital passports \cite{CompareAlg}. With rapid growth of usage of fingerprint-based systems, the size of fingerprint enrollment databases grow significantly. Therefore, efficient storage mechanisms are required to handle this massive amount of data. Also, in distributed biometric systems, sensors are physically located in a location different from the identification system itself. In most cases, fingerprint data are transmitted from sensor to processing system with low bandwidth and high latency wireless links. So, sensor data compression is necessary.

Two solutions can be offered to alleviate the data size problem. First, storing only the required features for identification (e.g., minutiae) instead of the raw data, which takes much less space as opposed to the raw data. Second solution is compressing the raw data. Since in some cases the algorithm may evolve over time, we may need to extract new features, which are not captured by the original feature set. Hence, the raw image is needed to perform enrollment phase again. Therefore, storing the compressed raw image is more desirable.

Compressing fingerprint data can be applied in two forms, Lossy or Lossless. Lossless compression can achieve up to 4:1 compression ratio, which may not yield sufficient storage size reduction \cite{CompareAlg}. The most widely used lossless compression formats include PNG, GIF, JPEG Lossless, JPEG-LS and JPEG2000 Lossless. Lossy compression can achieve desired compression rate for storing fingerprint image data. But, the compression may introduce distortion on the fingerprint images. This distortion can affect AFIS identification rate and decrease their performances. Compression distortion is usually measured by peak signal-to-noise ratio (PSNR) metric (i.e., the higher the PSNR, the higher quality of the compression). However, PSNR may not directly translate into the best quality metric for AFIS system as different distortions may affect the extracted minutiae information differently. Therefore, traditional quality metric are not used on fingerprint data; instead of them, measures that are obtained from end-to-end matching and identification results of AFIS are often used. These metrics include False Acceptance Rate (FAR), False Rejection Rate (FRR) and Equal Error Rate (EER) are used \cite{SurveyMatch}.

Although standards recommend specific compression algorithms to apply on fingerprint data, there exists other specialized algorithms targeting these type of images. Some of these algorithms used the special structure that exists in fingerprint images for compression. In \cite{Ridge} ridge structure is considered for compressing images. By focusing on similarity between structures bounded in small patches, \cite{Sparse} used Compressed Sensing (CS) for compression. This method requires training phase and considering the amount of fingerprint images in large datasets, performance of the algorithm might be reduced. In this paper, an algorithm proposed by focusing on special structure called \PSML\ (\psml). \psml\ refers to a structure that contains multiple parallel lines and could be drawn as a parametric geometric model. So, in comparison to CS methods, this kind of representation does not require pre-trained dictionary for encoding and decoding. The proposed method is comparable with JPEG2000 in common range of compression ratio, But it has a better performance in low compression rates (higher quality) and very high compression rates (very low quality).

The rest of the paper is organized as follows. The related works are discussed in section 2. Before presenting the proposed algorithm, model and related parameters are described in section 4, we first review the base algorithm (i.e., adaptive geometrical wavelet) in section 3. Section 5 presents with the details of the parameters quantization and model computation, which also presents our fast model computation algorithm. Section 6 provides the result of experiments and highlights the advantages of the proposed algorithm with evaluations on standard databases. Section 7 concludes the paper and suggests future directions of this work.

\section{Previous works}
By developing the usage of biometric data in last two decades, several algorithms and standards were set up for compressing data in biometric systems. The most relevant one ISO/IEC 19794 standard was developed for Biometric Data Interchange Format, where Part 4 focused on fingerprint data. ISO/IEC 19794-4 allows fingerprint images data to be stored in lossy manner in JPEG\cite{JPEG}, wavelet transform/scalar quantization (WSQ)\cite{WSQ} and JPEG2000\cite{JP2K1,JP2K2} formats, where the later is recommended. Also \cite{1000ppi} proposed a corresponding specific JPEG2000 Part I profile for 1000 ppi fingerprint images.

While using the data formats specified by the ISO/IEC 19794-4 standard established in most applications, the lack of recognition accuracy in comparison with other compression algorithms has led to development of new algorithms or using the general purpose ones for compressing fingerprint images. In general, the algorithms are divided in two categories, lossless or lossy, according to type of compression. But, this paper summarizes the previous works into four categories, which are discussed in the following.

The first category includes researches that review, compare and analyze existing methods proposed for compressing fingerprint images, any type of images, or any type of data. In \cite{comparewsqjp2} a complete comparison between JPEG and WSQ is done that shows WSQ have a better compression ratio against JPEG for fingerprint images. Based on the strong texture structure in fingerprint images, \cite{comparefractal} preformed a comparison between Wavelet Transform and Fractal Coding methods for texture-based images. Compression methods could be used together as a hybrid one, so \cite{comparehybrid} compared standard types of Wavelet, Fractal and JPEG. In addition, a comparison among standard types and hybrid types was performed. Later, due to the need of higher compression ratio for fingerprint images, \cite{comparelossy} performed a comparison among JPEG, JPEG2000, EZW and WSQ at higher compression ratio (40:1) and concluded that JPEG2000 has better performance than WSQ in higher compression ratio. This result was confirmed by \cite{NISTCompare} with PSNR and Spectral Image Validation and Verification (SIVV) \cite{NISTSIVV} metrics. In addition to above methods, \cite{CompareAlg} used set partitioning in hierarchical trees (SPIHT) and predictive residual vector quantization (PRVQ) compression methods in comparison. A fully comparison between lossless compression methods is done in \cite{comparelossless} that includes known methods like Lossless JPEG, JPEG-LS, Lossless JPEG2000, SPIHT, PNG, GIF and global compression methods like Huffman Coding, GZ, BZ2, 7z, RAR, UHA and ZIP. The metrics used for this comparison was FAR and FRR.

As stated above, methods based on Discrete Cosine Transform (DCT) and Wavelet Transform have most usage on fingerprint image compression. Also, coefficients related to higher frequency have more significance in fingerprint images rather than public images. Therefore, second category includes researches that focus on energy and coefficient distribution on these transforms. \cite{WaveSA,Psycho} used Simulated Annealing (SA) to optimize wavelets for fingerprint image compression by searching the space of wavelet filter coefficients to minimize an error metric. The distortion metric used by \cite{Psycho} is a combination of objective RMS error on compressed images and subjective evaluations by fingerprint experts. \cite{WaveletGA,WaveGABetter} proposed a similar approach by using Genetic Algorithm (GA) instead of SA. In \cite{WaveGABetter} fingerprint image divided into small patches (e.g. 32x32), then used GA to optimize the wavelet coefficients over them. Their evaluations show that the learning time was decreased while PSNR increased. Kasaei and et al. \cite{KasaeiTencon,KasaeiICIP,KasaeiIP} used Piecewise-Uniform Pyramid Lattice Vector Quantization to propose a method based on generalized Gaussian distribution for the distribution of the wavelet coefficient. Different filter banks that could be used in wavelet transform were reviewed in \cite{WaveFilter} and the conclusion is that using Coif~5 filter bank is preferred to Bior~7.9 filter bank, that is used in standard wavelet transform, producing higher compression ratio in fingerprint images. As an alternative strategy, special energy distribution based on fingerprint pattern can be used for determining coefficients. \cite{DCTDistrib} used this strategy for DCT-based coders.

The third category includes researches focused on structures founded in fingerprint images and used these structures for compression. The most prominent structure that can be found, is \textit{ridge}. Another structure-based compression is \textit{valley}. \cite{CompressRidgelet} used Ridgelet Transform and \cite{ValleySkel} used hybrid model from ridge and valley for compressing fingerprint images. The existence of these strong structures has led researchers to use Sparse Representation (SR). For instance \cite{SomLvq,SomWad} uses Self Organizing Map (SOM) -- is kind of SR -- to represent patches of fingerprint images in a compressed format. \cite{SomLvq} used SOM and Learning Vector Quantization (LVQ) to learn the network, while \cite{SomWad} used Wave Atom Decomposition and SOM. Also, k-SVD \cite{Sparse} is used for sparse representation of image patches. This method could achieve better performance against JPEG2000 and WSQ with PSNR metric.

The last category includes researches that could not be categorized in other categories. These researches provide new methods for compressing fingerprint images. In \cite{DfbTcq} a method is proposed based on Directional Filter Banks and Trellis Coded Quantization (TCQ). This method takes Wavelet-based Contourlet Transform from fingerprint image, then using TCQ, encodes the resulting coefficients and produces a compressed result. The results showed an improvement in performance over SPIHT. As mentioned before, lossless compression can also be used for compression. \cite{LosslessCompression} proposed a lossless compression method by changing the prediction function. The fingerprint image was divided into four parts, then the prediction function of top-right part changed to use information provided by itself and mirrored top-left part.

\section{Adaptive Geometrical Wavelet}
From human vision system, it is known that the human eye is designed to catch changes of location, scale and orientation \cite{NatureVision}. Classical wavelets were efficient in catching location and scale, but not efficient in catching orientation. Despite development of classical wavelet into 2D space \cite{WaveletTour,WaveletFriend,BeyondWavelets}, for efficient catching of orientation, Geometrical Wavelet was proposed. Geometrical wavelet can be divided into two groups, nonadaptive and adaptive. The first group is based on nonadaptive methods of computing, which use frames like Brushlets \cite{Brushlets}, Ridgelets \cite{Ridgelets}, Curvelets \cite{Curvelets}, Contourlets \cite{Contourlet} and Shearlets \cite{Shearlets}. The second group approximates image in an adaptive way. Most of them are based on dictionary, like Wedgelets \cite{Wedgelets}, Beamlets \cite{BeamletPyramid}, Second-order Wedgelets \cite{SecondWedge}, Platelets \cite{PlateletMed,PlateletPhd} and Surflets \cite{Surflets}. Recently, some approaches based on basis were proposed, like Bandelets \cite{Bandelets}, Grouplets \cite{Grouplets} and Tetrolets \cite{Tetrolet}.

In remaining of this section, basis definitions common to all adaptive geometrical wavelets are expressed. Then, the most known member of this family, Wedgelets, is briefly introduced.


\subsection{The Class of Horizon Functions}
\begin{figure}[!t]
\centering
\includegraphics[width=0.45\columnwidthe]{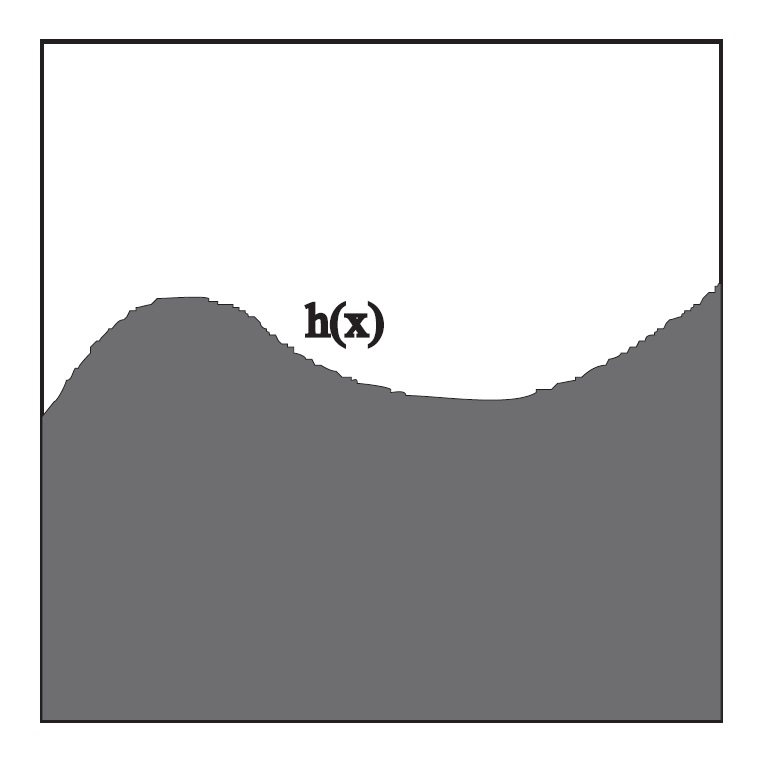}
\caption{An example of horizon and horizon function \cite{Wedgelets}}
\label{fig:Horiz}
\end{figure}

Let us define image domain $S=[0,1]\times[0,1]$. Function $h(x)$ defined in $S$ is called \textit{horizon}, if it is continuous, smooth, defined on interval $[0,1]$ and it fulfills the \Holder\ regularity condition. In practice, it is sufficient $h$ be a member of $C^2$ class. Assume characteristic function
\[H(x_1,x_2)=\mathbf{1}_{\{x_2 \le h(x_1)\}} \quad 0 \le x_1,x_2 \le 1.\]
Function $H$ is called a \textit{horizon function}, if $h$ is horizon. $H$ formed a black and white image that under the horizon is black and upper it is white. An example of horizon and horizon function is shown in Fig. \ref{fig:Horiz}.

\subsection{Dictionary of Wedgelets}
\begin{figure}[!t]
\centering
\includegraphics[width=0.6\columnwidthe]{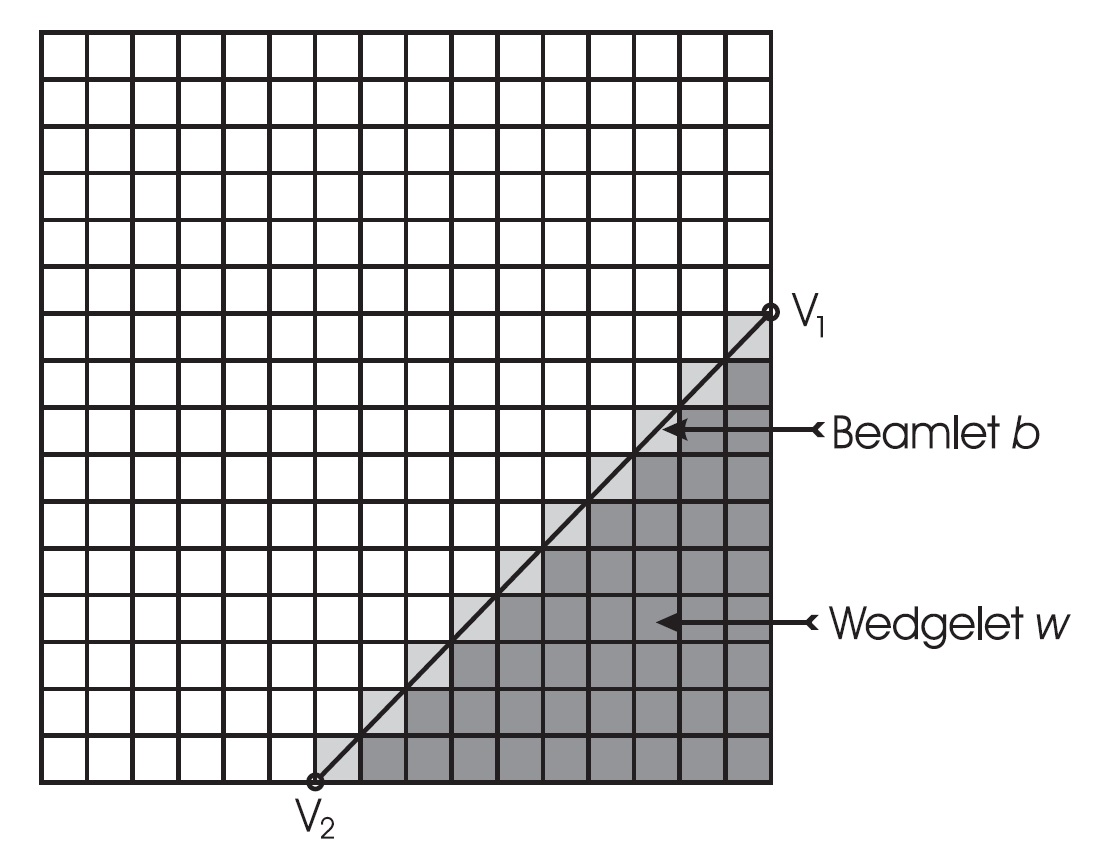}
\caption{An example of wedgelet defined by beamlet $b$ \cite{Wedgelets}}
\label{fig:BeamWedge}
\end{figure}

Denote dyadic square $S(j-1,j_2,i)$ defined as 2D range
\[ S(j_1,j_2,i)=[j_1/2^i,(j_1+1)/2^i]\times[j_2/2^i,(j_2+1)/2^i]\]
that $0 \le j_1,j_2 < 2^i \;;\; i \ge 0 \;;\; j_1,j_2,i \in \mathbb{N}$. Let's define image with $N\times N$ pixels. If assume $N=2^I$ then $S(0,0,0)$ denote the whole image domain and $S(j_1,j_2,I)$ denote all pixels of image. In each border of square $S(j_1,j_2,i)$ there exist vertices with distance equal to $\frac{1}{N}$. Every pair of such vertices can be connected to form a straight line -- edge (also called \textit{beamlet} after work of Donoho and Hou \cite{Wedgelets}). Therefore, one can define edges in different location, scale and orientation. The set of these edges forms a binary dictionary. Each edge $b$ takes square $S$ into two parts. Let us define one of the parts that is bounded between edge and up right corner with following indicator function
\[ w(x_1,x_2)=\mathbf{1}_{\{x_2 \le b(x_1)\}}\]
Such function is called \textit{wedgelet}, and is defined by beamlet $b$. The graphical representation of wedgelet defined by beamlet $b$ is shown in Fig. \ref{fig:BeamWedge}.

The set of wedgelets in any $S$ can be defined as 
\[ W(S)=\{\mathbf{1}(S)\} \cup \{\text{all possible}\ w\ \text{defined on}\ S\}.\]
From now, assume the subscripts $j_1,j_2$ such that $0 \le j_1,j_2 < 2^i$ is replaced by subscript $j$ such that $0 \le j < 4^i$. Also lets denote square $S(j_1,j_2,i)$ as $S_{i,j}$. Also, lets denote parametrization of orientation with $m$, which was denoted coordinate $v_1,v_2$ previously.
\begin{definition}
The \textit{Wedgelets Dictionary} are defined as the following set
\begin{multline}W=\{w_{i,j,m} : i=0,\cdots,\log_2 N \;;\; j=0,\cdots,\\4^i-1 \;;\; m=0,\cdots,M_W(S_{i,j})-1 \}\end{multline}
where $M_W(S_{i,j})$ denotes the number of wedgelets in $S_{i,j}$.
\end{definition}

\subsection{Wedgelet Transform}
Consider image as $F:S\rightarrow\mathbb{N}$, so the wedgelets transform is defined as following
\begin{definition}
\textit{Wedgelets Transform} defined as following formula
\[ \alpha_{i,j,m}=\frac{1}{T}\int\int_S F(x_1,x_2)w_{i,j,m}(x_1,x_2)\mathrm{d}x_1\mathrm{d}x_2 \]
where
\[ T=\int\int_S w_{i,j,m}(x_1,x_2)\mathrm{d}x_1\mathrm{d}x_2\]
is the normalization factor and $S=[0,1]\times[0,1]$, $\alpha_{i,j,m}\in\mathbb{R}$, $w_{i,j,m}\in W$, $0 \le i \le log_2 N$ , $0 \le j < 4^i$, $0 \le m < M_W(S_{i,j})$ and $i,j,m \in\mathbb{R}$.
\end{definition}
In grayscale images, coefficients quantized to $\alpha_{i,j,m}\in \{0,\cdots,255\}$, while in binary images coefficients are quantized to $\alpha_{i,j,m}\in\{0,1\}$. Finally, image representation with wedgelets is defined as following:
\[F(x_1,x_2)=\sum_{i,j,m}\alpha_{i,j,m}w_{i,j,m}(x_1,x_2)\]
But, because $W$ is dictionary, not basis, not all of coefficients in above formula are used \cite{SecondWedge} (It means some of coefficient are set to zero). At last, we are looking for the best approximation image using minimum number of atoms from a given dictionary.

\subsection{Wedgelet Analysis of Image}
Almost all multiresolution methods, including wedgelet approximation, use quadtree as main data structure. Although many ways exist to store wedgelets with the help of quadtree, but the most common way is the one that assumes in each node of quadtree, the coefficients determining the appropriate wedgelet are stored.

The basic algorithm of decomposition image, contains two steps. In first step, full image decomposition using wedgelets transform is obtained. It means. for each square $S$, the best approximation with minimum square error using wedgelet is obtained. In second step, a bottom-up optimization pruning algorithm is applied on decomposition tree to obtain best quality approximation image with minimum number of atoms. Indeed, the following weighted formula is minimized \cite{Wedgelets}:
\[R_{\lambda}=\min_{P}\{\parallel F-F_W\parallel_2^2+\lambda^2 K\},\]
where $P$ is homogeneous partition of an image, $F$ denotes original image, $F_W$ is wedgelet approximation, $K$ is the number of bits required to encode approximation and $\lambda$ is rate distortion parameter.
\section{Proposed method}
Fingerprint images can be viewed simply as black and white images. In this form of images, fingerprint images seem to be made of by black lines mashing around themselves in white background. If a small patch of image is considered, a couple of parallel black lines are seen. So, this aspect of view, represents a geometric model that is built up with parallel lines with arbitrary widths, named \textit{parallel stroked multi lines} model. In this paper, an algorithm based on Adaptive Geometrical Wavelet is used to employ \psml\ model for compressing fingerprint images. in the rest of this section, the proposed model and its parameters are described in details.

\subsection{\PSML\ model}
\begin{figure}[!t]
	\centering
	\includegraphics[width=0.5\columnwidthe,trim=5.2cm 15cm 5.5cm 4.3cm,clip]{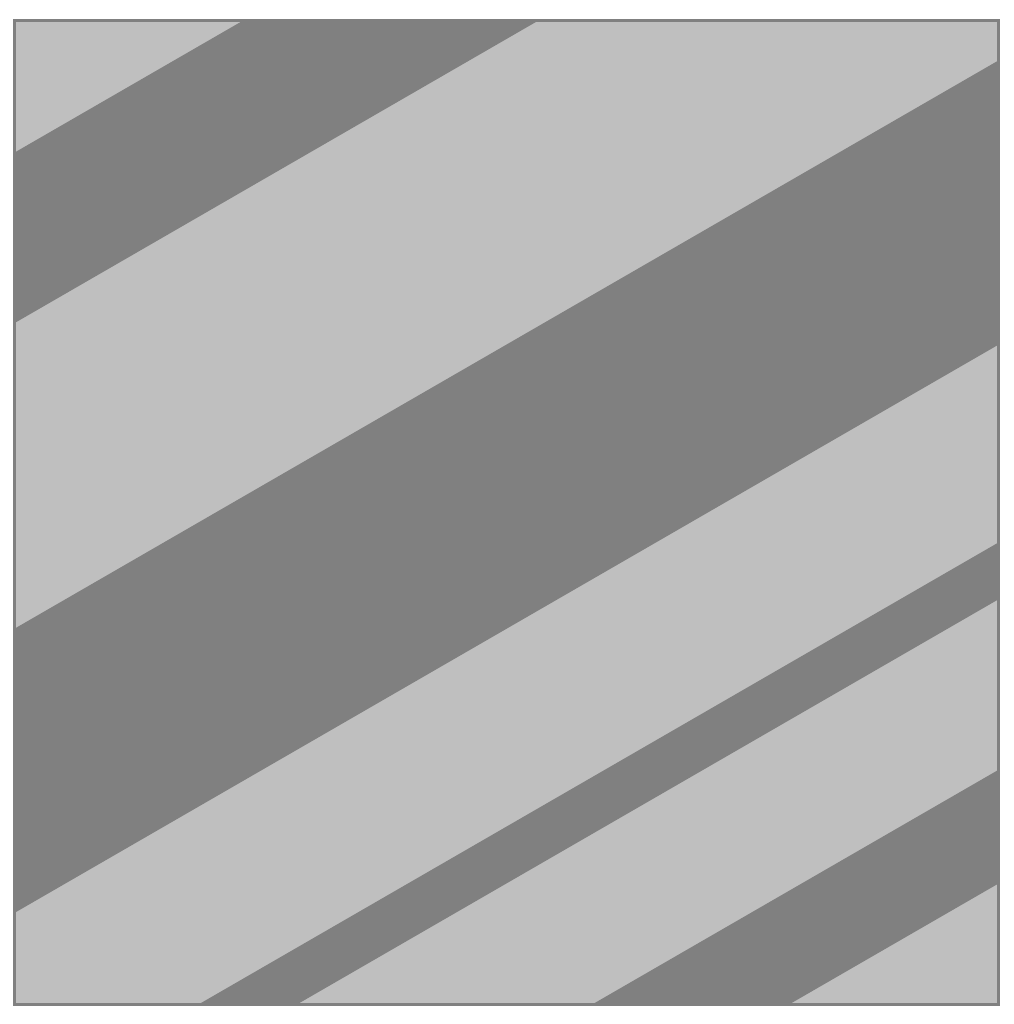}
	\caption{An example of the \psml\ model}
	\label{fig:psml1}
\end{figure}

\PSML\ model is built up by some parallel lines. Each line may have an arbitrary width. The gray level value of all pixels lie on each line is the same. Likewise, the gray level value of all background pixels, between lines, are the same (see Fig. \ref{fig:psml1}). 
\begin{definition}
Assume rectangle $S$ with arbitrary size and parameters $\theta$, $k$, $c_1$ and $c_2$. Suppose part of $k$ lines $l_1$ to $l_k$ with width $w_{l_1}$ to $w_{l_k}$ respectively and slope $\theta$ that lies in $S$. By considering $c_1$ as gray level value of lines and $c_2$ as the value of $S$ gray level, rectangle $S$ defines a gray level image. This image forms \PSML\ model.
\end{definition}

\subsection{Model parameters}
For simplicity of model representation, each line $l_i$ with width $w_{l_i}$ is assumed as $w_{l_i}$ lines with unit width. The background part between each two lines ($l_i$ and $l_{i+1}$) also assumed as some lines with unit width. Thus, the proposed model can be represented by a set of parallel lines with unit width, slope $\theta$ and gray level value $c_1$ or $c_2$. Considering the rectangle $S$ is made by $p$ lines with unit width and slope $\theta$, then the gray level value of all lines in $S$ can be represented by binary stream $G=g_1 g_2 \cdots g_p$ (zero means $c_1$ and one means $c_2$). Therefore, \psml\ can be represented with parameters $\theta$, $c_1$, $c_2$ and $G$.

\section{Model Computation}
To compute the model, first parameter quantization is described. After that, time complexity of computation is analyzed, then, an approximation method is proposed to speed up the proposed algorithm. Finally, a simple compression algorithm proposed to compress the resulting coefficients by the \psml\ algorithm.

\subsection{Model parameters quantization}
In this section, parameter quantization is described, separately. Parameter $\theta$ varies in range $[0,\pi)$ to cover all directions. For this purpose two points $C$ and $O$ were used to represent orientation. The angle between the line passing from points $C$ and $O$ and the horizontal axis of coordinate system, forms $\theta$. Suppose that square $S$ is $m\times n$. Point $C$ is fixed and located at pixel $C=(\lfloor \frac{m}{2} \rfloor,\lceil \frac{n}{2} \rceil)$. Location of point $O$ varies and can be any pixel in set $D_O=\{(0,y)|y=0,\ldots,n-1\}\cup\{(x,n-1)|x=0,\ldots,m-1\}$. Thus, the number of difference state of parameter $\theta$ is equal to $m+n-1$.

The value of gray level parameters $c_1$ and $c_2$ are usually represented by 8 bits. But, if needed, it can be reduced for better compression. Finally, binary stream $G$ is coded maximumly with $\sqrt{m^2+n^2}$ bits.

\subsection{Time complexity of model computation}
\begin{theorem}
The time complexity of approximating $n\times n$ patch with \PSML\ and using MSE metric is $O(n^3 2^n)$.
\end{theorem}
\begin{thmproof}
Each patch with size $m\times n$ has $(m+n-1)2^{\sqrt{m^2+n^2}}$ different structural states. For $n\times n$ patch approximately $O(n 2^n)$ different structural states exist. In each of these states, the best value for $c_1$ (or $c_2$) is computed by mean of gray level of pixels that have value zero (or one) in binary stream $G$. So, to calculate the best approximation of $n \times n$ patch, all of these states should be computed and the one with lowest MSE is chosen. The calculating the mean values in each patch needs $O(n^2)$ operations. Therefore, best approximation of a patch has $O(n^3 2^n)$ time complexity.
\end{thmproof}

\subsection{Fast patch approximation algorithm}
Since the time complexity of approximation of each patch has exponential order, an approximation method is proposed to compute it faster. The proposed method is based on \textit{Hill Climbing} algorithm and finds the local maximum solution. In this method, parameter $\theta$ is known and we try to find parameter $G$. After this parameter is computed, parameters $c_1$ and $c_2$ could easily be computed.

\begin{figure}[!t]
\centering
\includegraphics[width=0.95\columnwidthe]{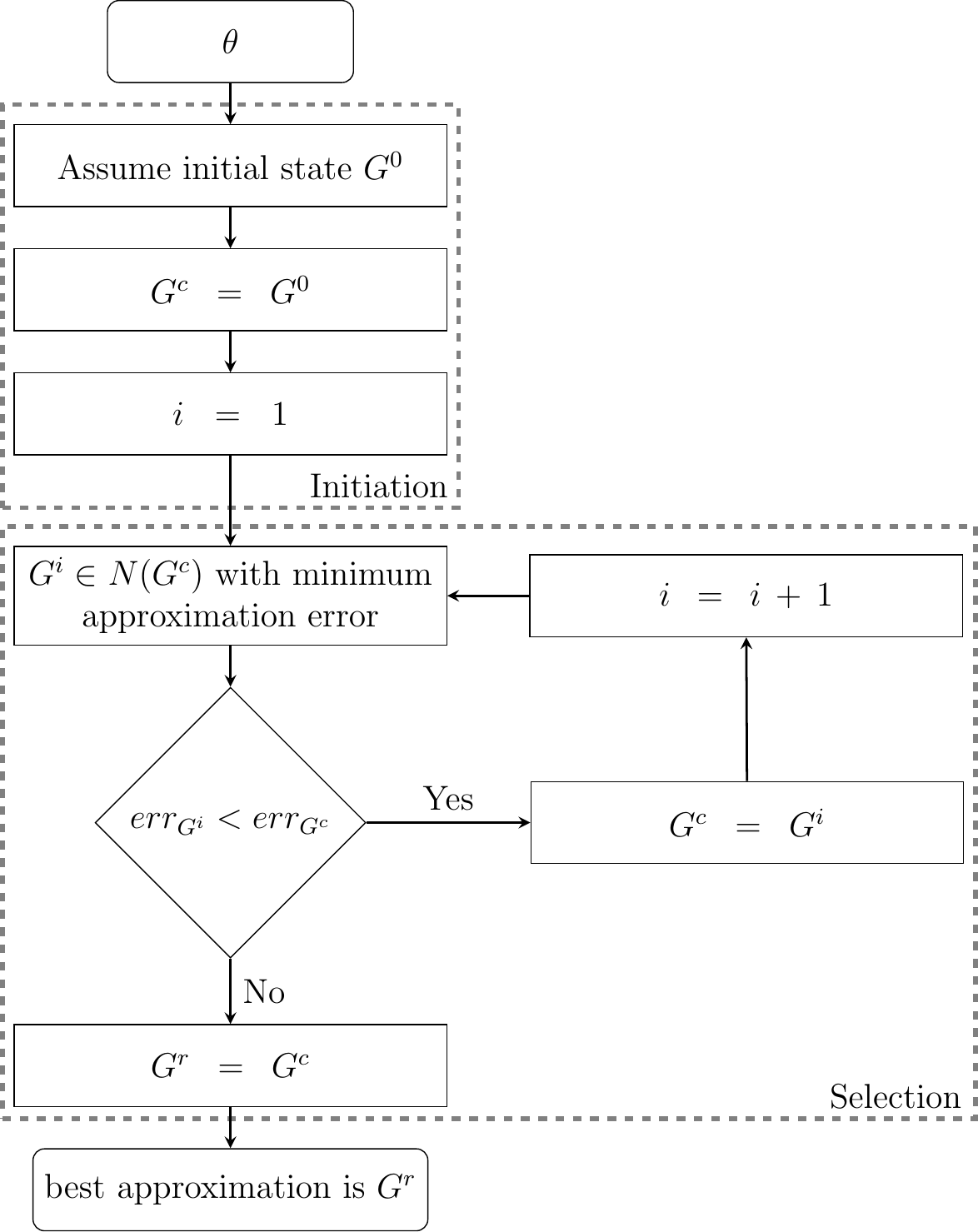}
\caption{The flowchart of Algorithm \ref{alg:fastapprox} (Fast Approximation) that compute patch parameters with fast approximation algorithm.}
\label{fig:fastapprox}
\end{figure}

\begin{algorithm}[Fast Approximation]
\label{alg:fastapprox}

\textbf{Definition}: Denote binary stream $G=g_1 g_2 \cdots g_p$. Every binary stream $G_k$ ($i=1,\cdots,p$) is defined as
\[ G_k=g'_1 g'_2 \cdots g'_p \quad\left\{\begin{array}{ll} g'_j=g_j&j\neq k \\ g'_j=\text{not}(g_j)& j=k\end{array}\right.\]
and is called \textit{neighbors of binary stream $G$}. Let's denote all neighbors of state $G$ with $N(G)$.

\textbf{Initiation}: Assume initial state as $G^0$ and current state as $G^c$. To initialize the algorithm set $G^c$=$G^0$ and $i=1$.

\textbf{Selection}: In each step $i$ of algorithm, the neighbor $G^i \in N(G^c)$ with lower approximation error than all states in $N(G^c)$, is selected. If $err_{G^i}<err_{G^c}$, set $G^c=G^i$ and do the selection step again with $i=i+1$. If $err_{G^i} \ge err_{G^c}$ then $G^r=G^i$ and go to the next step.

\textbf{Output}: The best approximation with lowest error of patch with known parameter $\theta$ is $G^r$.
\label{alg:fast}
\end{algorithm}
The flowchart of this algorithm is depicted in Fig. \ref{fig:fastapprox}.

From the experiments, the best initial state that produces result near the optimal, is generated by the following algorithm.
\begin{algorithm}[Initial State]
Denote the mean of all gray level value of all pixels in the image as $c_m$. For each line $l_i (i=1,\cdots,p)$, If the mean of gray level value of all pixels of the line is lower than $c_m$, then let $g_i=1$, else $g_i=0$. The produced binary stream forms the \textit{initial state}.
\end{algorithm}
\begin{corollary}\label{col:seltime}
From the experimental results, the selection step of algorithm \ref{alg:fast} is repeated in order of $O(n)$.
\end{corollary}
\begin{lemma}
The best approximation of each $n\times n$ patch with \psml\ and fast approximation algorithm is computed in $O(n^5)$.
\end{lemma}
\begin{thmproof}
The approximation algorithm for each $n\times n$ patch is computed with following algorithm:
\begin{center}
\begin{tabular}{lll}
1& Do for each orientation $\theta$ &$O(n)$ \\
2&\quad\; Compute initial state &$O(n^2)$ \\
3&\quad\; Repeat the selection phase &$O(n)$ (corollary \ref{col:seltime}) \\
4&\quad\;\quad\; Do for each neighbor in $N(G^i)$&$O(n)$ \\
5&\quad\;\quad\;\quad\; Calculate $c_1$ and $c_2$ parameters for $G_k$&$O(n^2)$
\end{tabular}
\end{center}
Accordingly, the best approximation of each $n \times n$ patch is computed in $O(n^5)$.
\end{thmproof}

Let's suppose each line with unit width in rectangle $S$ is denoted by $u_i$ ($i=1,\ldots,p$). Sum of pixel gray level values and number of pixels belonging to each line are denoted by $N_i$ and $S_i$, respectively. Then, the $c_1$ and $c_2$ parameters are calculated as
\[ c_1=\frac{\sum_{\substack{i=1,\ldots,p\\g_i=0}}S_i}{\sum_{\substack{i=1,\ldots,p\\g_i=0}}N_i} \; , \; 
    c_2=\frac{\sum_{\substack{i=1,\ldots,p\\g_i=1}}S_i}{\sum_{\substack{i=1,\ldots,p\\g_i=1}}N_i}.\]
This form of computation reduces the time complexity of calculating $c_1$ and $c_2$ parameters to $O(n)$. Surprisingly, this time complexity can reduce much more according to the following lemma.
\begin{lemma} \label{lem:O1}
The time complexity of calculating $c_1$ and $c_2$ parameters for $G_k \in N(G)$ is $O(1)$.
\end{lemma}
\begin{thmproof}
Let's define $SC_{c,G}$ and $NC_{c,G}$ as
\[ SC_{c,G}=\sum_{\substack{i=1,\ldots,p\\g_i=c}}S_i \; , \; NC_{c,G}=\sum_{\substack{i=1,\ldots,p\\g_i=c}}N_i .\]
The parameters $c_1^k$ and $c_2^k$ of $G_k$ can be computed as
\[ \left\{\begin{array}{lll}\text{if}\; g_i=0 &\quad& c_1^k=\frac{SC_{0,G}-S_k}{NC_{0,G}-N_k} \;,\;  c_2^k=\frac{SC_{1,G}+S_k}{NC_{1,G}+N_k} \\ \text{if}\; g_i=1 &\quad& c_1^k=\frac{SC_{0,G}+S_k}{NC_{0,G}+N_k} \;,\;  c_2^k=\frac{SC_{1,G}-S_k}{NC_{1,G}-N_k}
\end{array}\right.,\]
that takes $O(1)$ time complexity. Also, the parameters $SC_{c,G^k}$ and $NC_{c,G^k}$ can be computed in $O(1)$.
\end{thmproof}
By using the lemma \ref{lem:O1}, the approximation algorithm for each $n \times n$ patch is computed with following algorithm:
\begin{center}
\begin{tabular}{lll}
1& Do for each orientation $\theta$ &$O(n)$ \\
2&\quad\; Compute initial state ($S_i,N_i,SC_{c,G^0},NC_{c,G^0}$) &$O(n^2)$ \\
3&\quad\; Repeat the selection phase &$O(n)$ (corollary \ref{col:seltime}) \\
4&\quad\;\quad\; Do for each neighbor in $N(G^i)$&$O(n)$ \\
5&\quad\;\quad\;\quad\; Calculate $c_1$ and $c_2$ parameters for $G_k$&$O(1)$ (lemma \ref{lem:O1})
\end{tabular}
\end{center}
\begin{theorem}
Time complexity of the fast approximation algorithm used to obtain the best approximation of the whole $n \times n$ image with \PSML\ is $O(n^3)$.
\end{theorem}
\begin{thmproof}
Based on lemma \ref{lem:O1}, the best approximation of each $m \times m$ patch is computed in $O(m^3)$. Without loss of generality, we assume that $n$ is dyadic ($n=2^J$). The proposed method is based on the adaptive geometrical wavelet. According to \cite{Wedgelets}, when using quadtree for partitioning image, time complexity for the whole image, $\Psi(n)$, is computed as bellow: 
\[
\begin{array}{ll}
\Psi(n)&=\sum_{j=0}^{J} 2^{2j} \times a \times (2^{J-j})^3 \\&=a\sum_{j=0}^{J}2^{2J}2^{(J-j)} = a 2^{2J}\sum_{j=0}^{J} 2^{j} \\
&= a n^2(2^{J+1}-1) = a(2n^3-n^2) \\
&=O(n^3),
\end{array}
\]
where $a$ is a constant.
\end{thmproof}

\subsection{Model compression}
\label{sec:modelcompress}
Most compression algorithms like JPEG2000 use complex relationship among pixels to predict gray level value of a pixel. Thus, the redundancy between gray level values is reduced and the compression algorithms achieve lower bit rate. To have a fair comparison between the \psml\ algorithm and other compression methods, the redundancy in \psml\ coefficients must be reduced. In the proposed method, the gray level and structural parameters of each pixels are similar to the neighbor patches. So, a compression algorithm can be applied on these parameters. In this paper, a simple compression algorithm based on Low Complexity Lossless Compression for Images (LOCO-I) predictor and Golomb-Rice coding \cite{golomb-rice} is used to compress just the gray level coefficients obtained from the \psml\ model.

\section{Experimental results}
At first part of this section, the methods, tools and databases used for experiments are described. Then in second part, the experiments and parameters used for each one of them are explained. Also, a complete discussion on results obtained by the experiments is provided.
\subsection{Settings and Methods}
\subsubsection{Compression methods}
Nearly all the new methods proposed for compressing fingerprint images, compare their works with JPEG2000 method. Therefore, for sake of consistency, the \psml\ algorithm is compared against the JPEG2000. In this paper, implementation provided by Kakadu Software \cite{kakadu} is used. In addition we compare the \psml\ algorithm against Wedgelets Transform proposed by \cite{Wedgelets}.

\subsubsection{Biometric recognition systems}
Among the well-known biometric recognition systems (e.g. VeriFinger, eFinger and NIST FIVB), we choose VeriFinger to use in this paper. VeriFinger is a commercial tool that acts based on minutiae matching. Experiments have shown that VeriFinger is more robust against noise and compression artifact compared to the other tools. VeriFinger feature matching algorithm provides similarity score as the result. The higher is the score, the higher is probability that feature collections are obtained from same person. The version used in this paper was VeriFinger SDK v7.1 published on 24/08/2015 \cite{verifinger}.

\subsubsection{Databases}
to evaluate the proposed method, three fingerprint images databases are used that shortly named DB1, DB2 and DB3. These databases are described in the following:

\textbf{DB1}: This database is \textit{DB1\_B} from \textit{Fingerprint Verification Competition (FVC2002)} that contains 10 fingers and 8 impressions per finger (80 fingerprints in all). Each fingerprint has $388\times 374$ pixels obtained with 500 dpi resolution. The image with name \textit{'101\_1.tif'} is called PIC1.

\textbf{DB2}: This database is \textit{DB4\_B} from \textit{Fingerprint Verification Competition (FVC2002)} that contains 10 fingers and 8 impressions per finger (80 fingerprints in all). Each fingerprint has $288\times 384$ pixels obtained with about 500 dpi resolution. The image with name \textit{'103\_1.tif'} is called PIC2.

\textbf{DB3}: This database contains fingerprint images from 7 persons. Each person has data for all 10 fingers (except one that has data for 5 fingers) and 8 impressions exist per finger (520 fingerprints overall). These fingerprint obtained by \textit{U.are.U 400} biometric hardware with $326 \times 357$ pixels, 500 ppi resolution and published by \textit{NeuroTechnology} company. Due to the fact that fingerprint data of each finger is different from other fingers in same person, they could be assumed as 65 fingers and 8 impression per finger. The image with name \textit{'012\_1\_2.tif'} is called PIC3.

\subsection{Experiments}
\begin{figure*}[!t]
\centering
\subfloat[]{\includegraphics[width=0.9\columnwidthe,trim=0.8cm 0.2cm 1cm 0cm,clip]{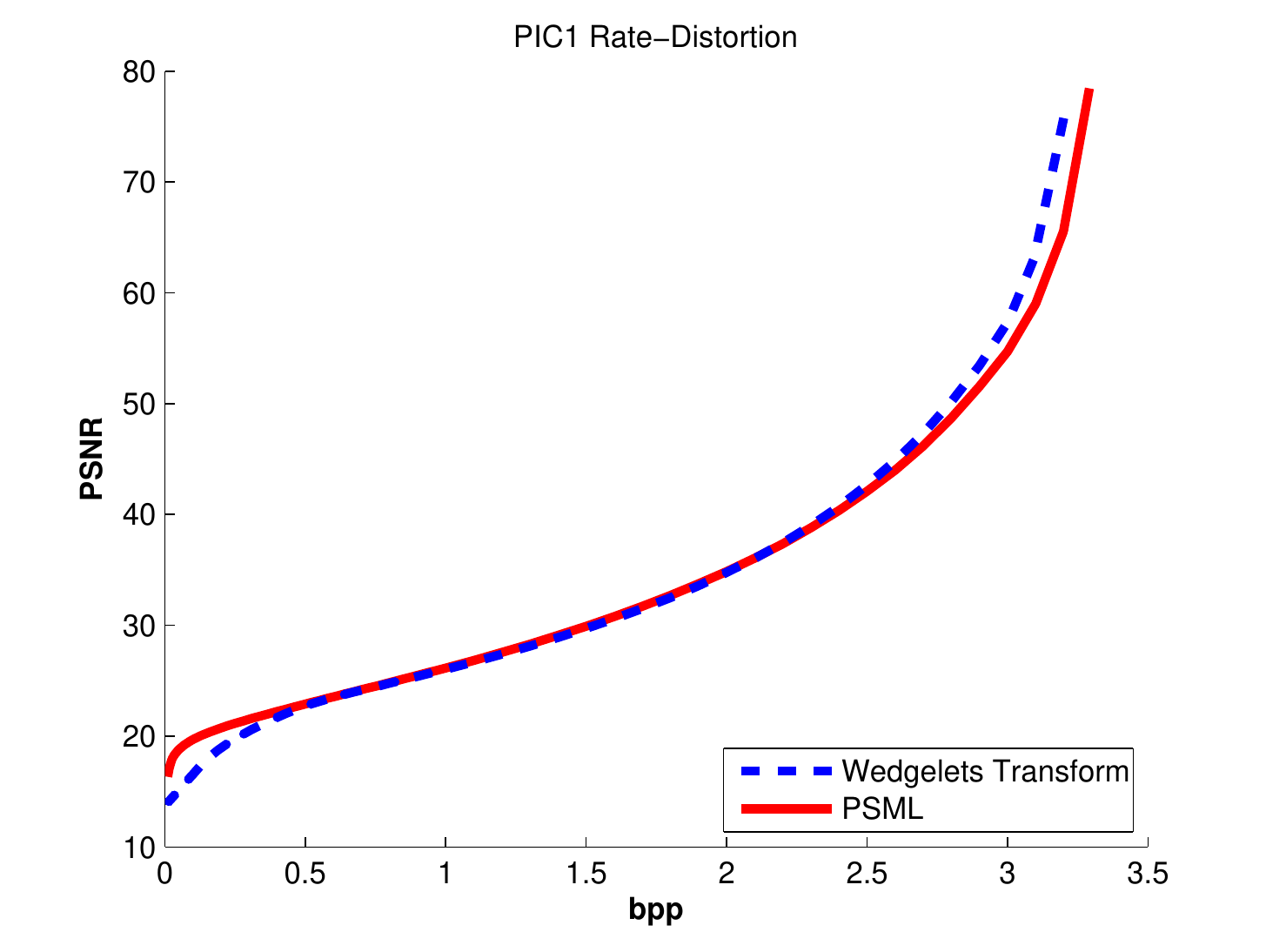}\label{fig:WMrdPIC1bpp}}
\hfill
\subfloat[]{\includegraphics[width=0.9\columnwidthe,trim=0.8cm 0.2cm 1cm 0cm,clip]{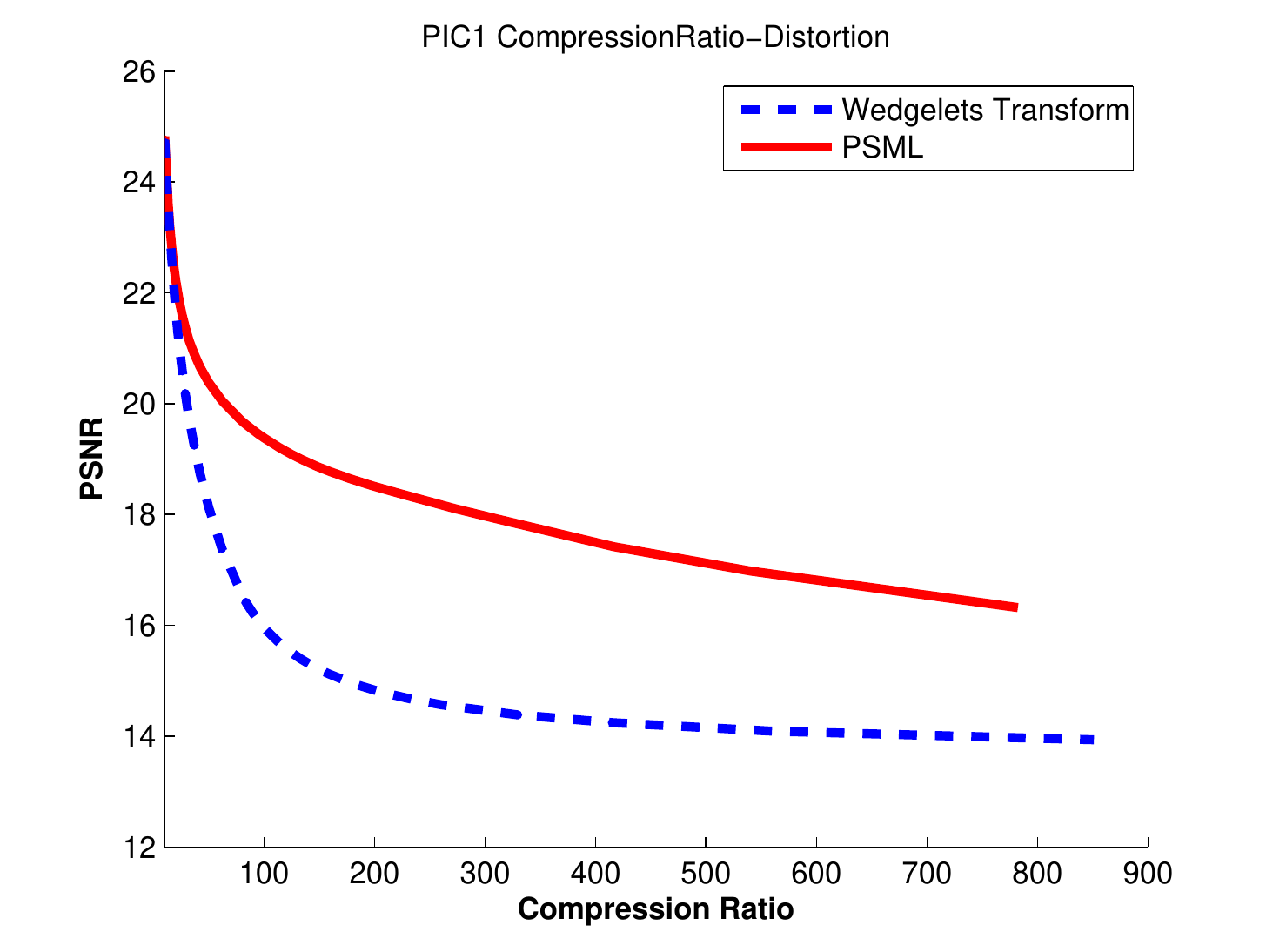}}
\\
\subfloat[]{\includegraphics[width=0.9\columnwidthe,trim=0.8cm 0.2cm 1cm 0cm,clip]{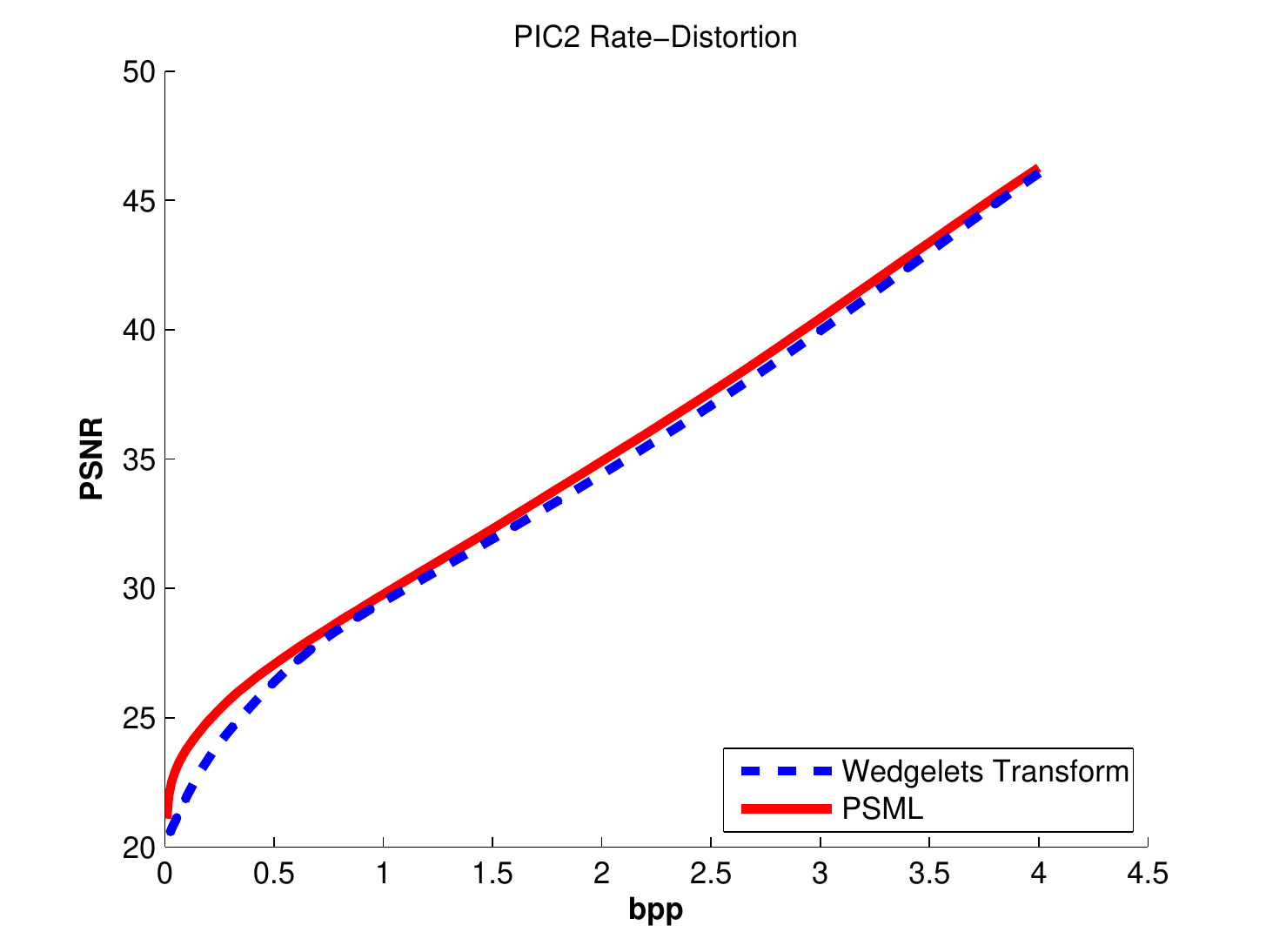}}
\hfill
\subfloat[]{\includegraphics[width=0.9\columnwidthe,trim=0.8cm 0.2cm 1cm 0cm,clip]{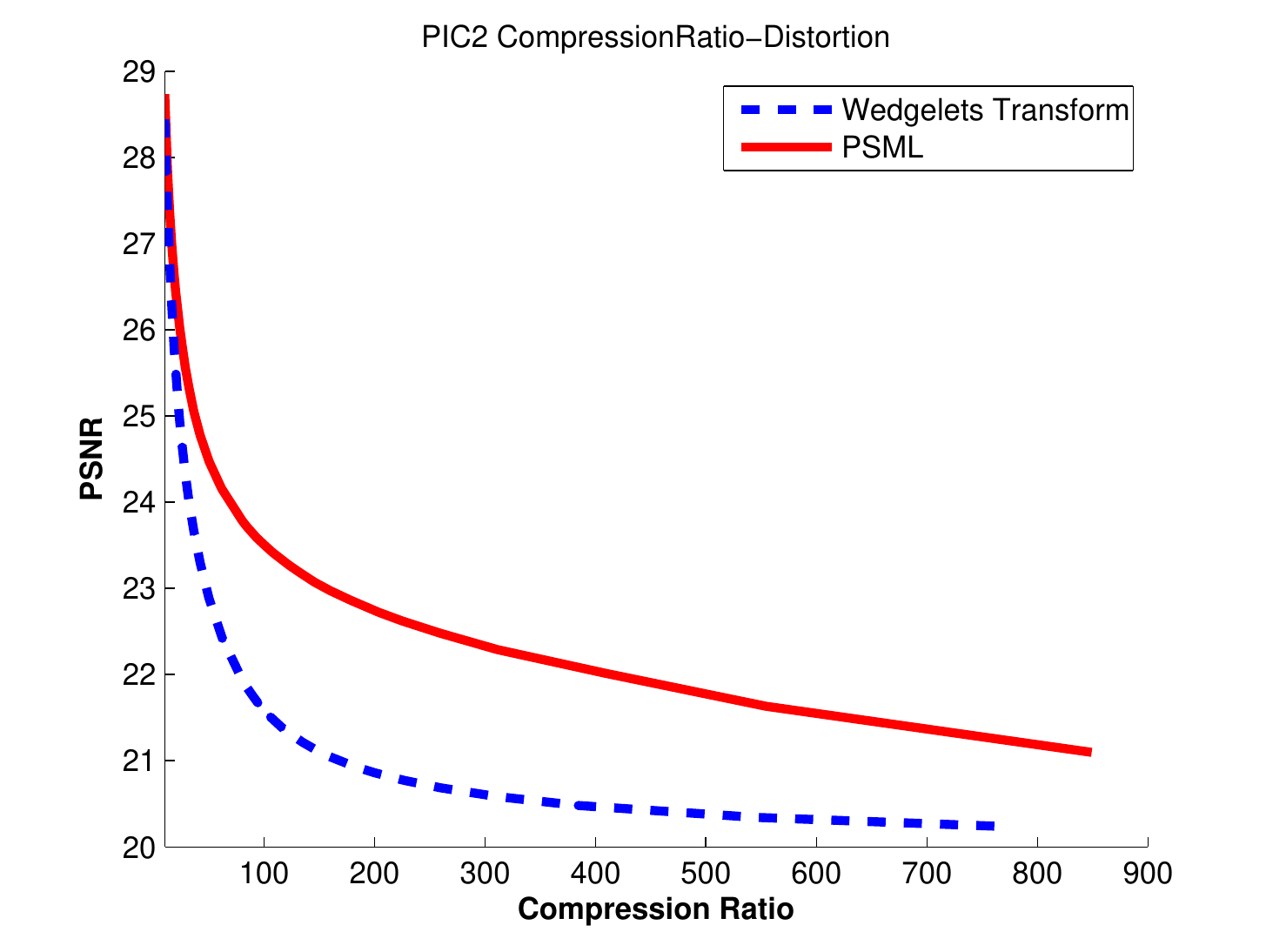}}
\\
\subfloat[]{\includegraphics[width=0.9\columnwidthe,trim=0.8cm 0.2cm 1cm 0cm,clip]{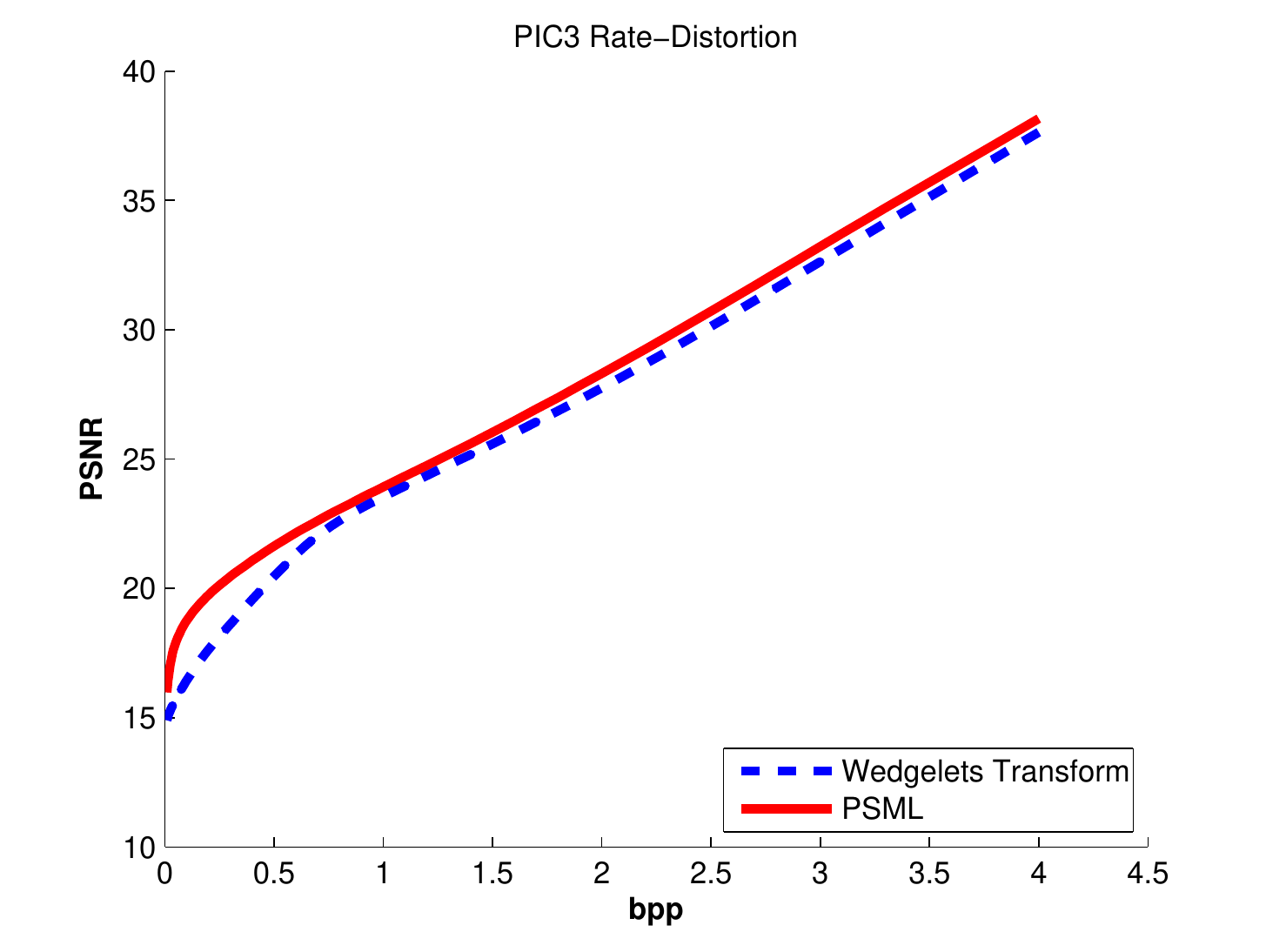}}
\hfill
\subfloat[]{\includegraphics[width=0.9\columnwidthe,trim=0.8cm 0.2cm 1cm 0cm,clip]{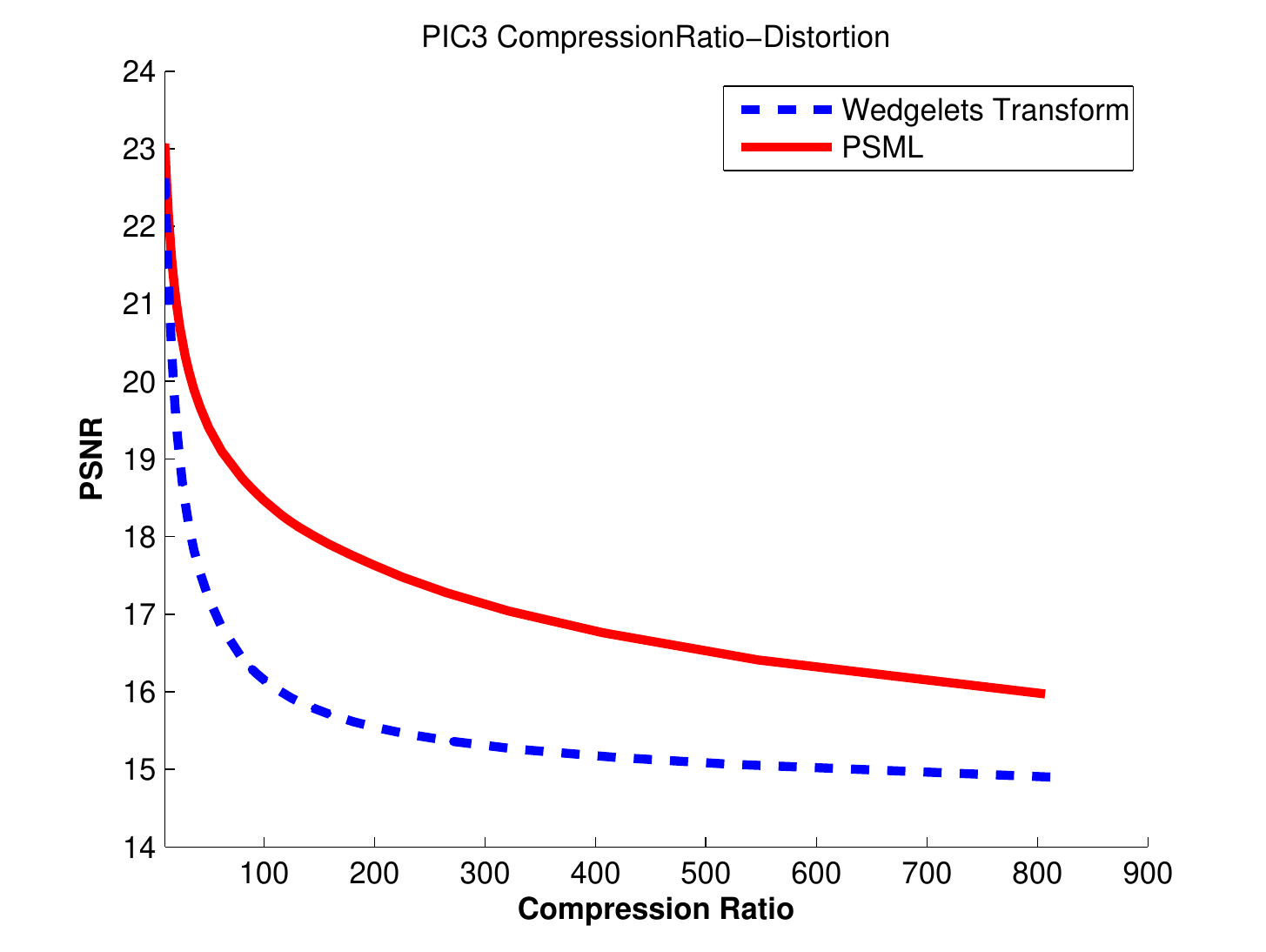}}
\caption{Compare rate-distortion (left) and compression ratio-distortion (right) for Wedgelets Transform and \psml\ algorithm according to experiment~1. Rate-distortion diagram shows results of low compression rates in detail, while compression ratio-distortion diagram shows result of high compression rates in detail. (a,b) PIC1 (c,d) PIC2 (e,f) PIC3.}
\label{fig:WMrd}
\end{figure*}

\begin{figure*}[!t]
\centering
\makebox[0.3\columnwidthe][c]{\tiny Original image} \quad \makebox[0.3\columnwidthe][c]{\tiny Wedgelets (1 bpp)} \quad \makebox[0.3\columnwidthe][c]{\tiny PSML (1 bpp)} \quad \makebox[0.3\columnwidthe][c]{\tiny Wedgelets (0.1 bpp)} \quad \makebox[0.3\columnwidthe][c]{\tiny PSML (0.1 bpp)}
\\
\subfloat[]{\includegraphics[width=0.3\columnwidthe]{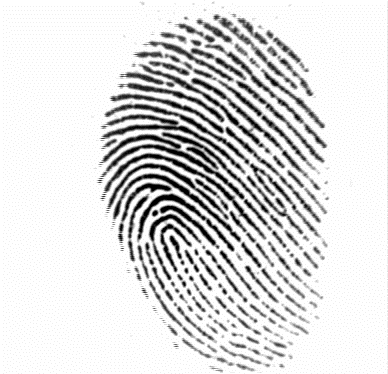}}
\quad
\subfloat[]{\includegraphics[width=0.3\columnwidthe]{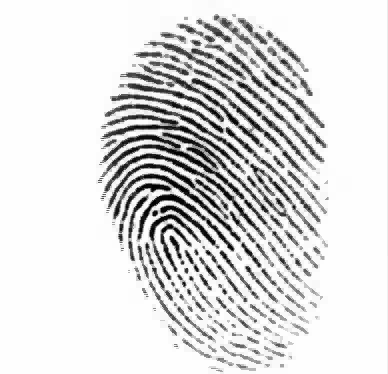}}
\quad
\subfloat[]{\includegraphics[width=0.3\columnwidthe]{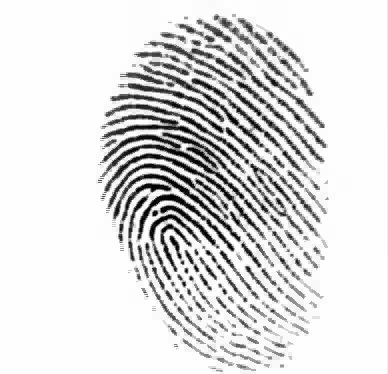}}
\quad
\subfloat[]{\includegraphics[width=0.3\columnwidthe]{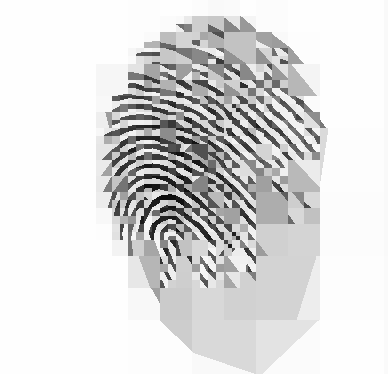}}
\quad
\subfloat[]{\includegraphics[width=0.3\columnwidthe]{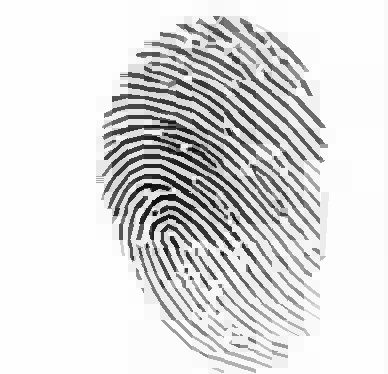}}
\\
\subfloat[]{\includegraphics[width=0.3\columnwidthe]{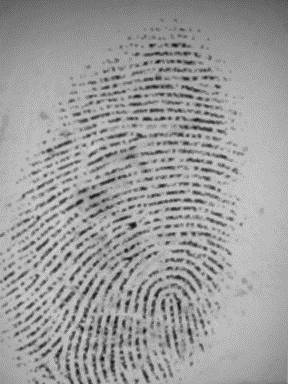}}
\quad
\subfloat[]{\includegraphics[width=0.3\columnwidthe]{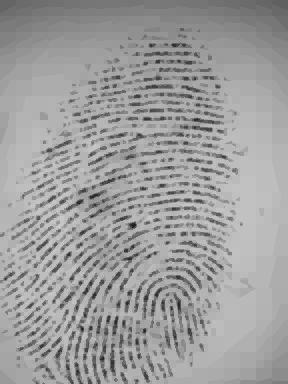}}
\quad
\subfloat[]{\includegraphics[width=0.3\columnwidthe]{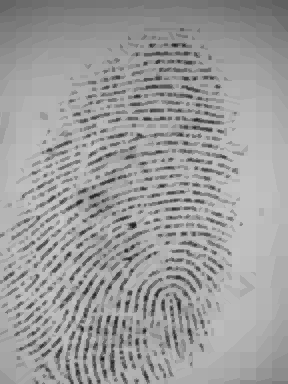}}
\quad
\subfloat[]{\includegraphics[width=0.3\columnwidthe]{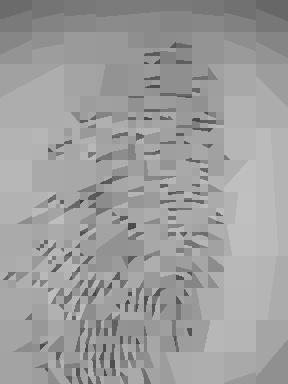}}
\quad
\subfloat[]{\includegraphics[width=0.3\columnwidthe]{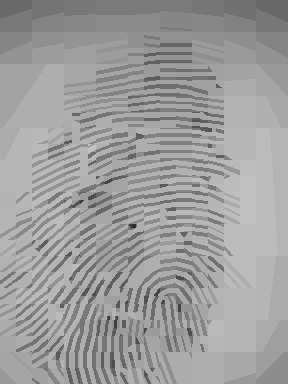}}
\\
\subfloat[]{\includegraphics[width=0.3\columnwidthe]{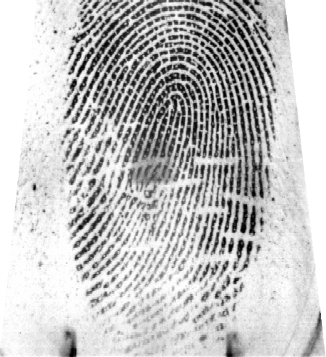}}
\quad
\subfloat[]{\includegraphics[width=0.3\columnwidthe]{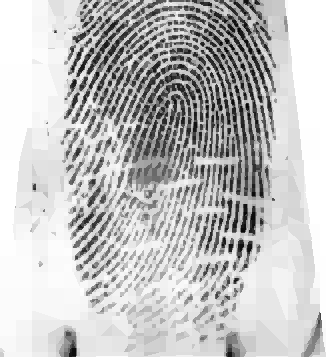}}
\quad
\subfloat[]{\includegraphics[width=0.3\columnwidthe]{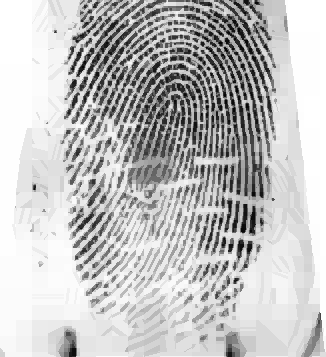}}
\quad
\subfloat[]{\includegraphics[width=0.3\columnwidthe]{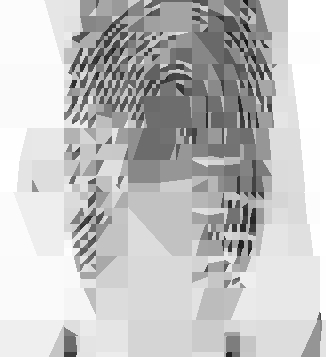}}
\quad
\subfloat[]{\includegraphics[width=0.3\columnwidthe]{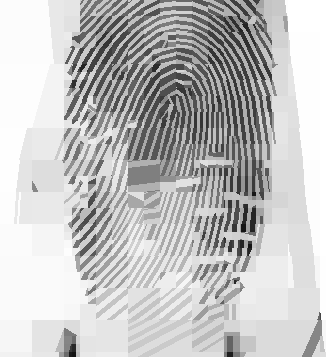}}
\caption{Samples of fingerprint images encoded by Wedgelets Transform and \psml\ algorithm, according to Experiment 1. First column shows the original fingerprint image. Second and fourth columns show the compressed fingerprint images by Wedgelets Transform. Third and Fifth columns show the compressed fingerprint images by \psml\ algorithm. Second and Third columns are encoded in 1 bpp, while fourth and fifth column are encoded in 0.1 bpp. First row belongs to result of experiment 1 on PIC1, while second and third rows belongs to PIC2 and PIC3, respectively. The PSNR value of images are (b,c) 26.03 , 26.15 (d,e) 16.51 , 19.69 (g,h) 29.49 , 29.78 (i,j) 21.94 , 23.78 (l,m) 23.54 , 23.92 (n,o) 16.43 ,18.76.}
\label{fig:WMpic}
\end{figure*}

\textbf{Experiment 1}: All images in DB1, DB2 and DB3 are encoded with Wedgelets Transform and \psml\ algorithm at different compression rates. The settings used for both algorithms are 8 bits for gray level value, builds quadtree up to 7 levels of model and do not use compression on the resulting coefficients discussed on \ref{sec:modelcompress}. According to the results, almost all images of all databases produced the same results across different rates with PSNR measure. Fig. \ref{fig:WMrd} shows compression result for PIC1, PIC2 and PIC3 obtained in compression rates from 0.01 to 4 bit per pixel (bpp). In this figure, two types of diagram are provided. The rate-distortion diagrams provide the better visualization of results in lower compression rates, while the compression ratio-distortion diagrams provide the better one in higher compression rates. In addition, Fig. \ref{fig:WMpic} shows the compressed images of PIC1, PIC2 and PIC3 at two different rates, 1 bpp and 0.1 bpp, with both compression algorithms.

%
%

\textbf{Corollary 1 (Wedgelets vs \psml)}: In experiment 1, both \psml\ algorithm and Wedgelets Transform used the same configurations. In lower compression rates, the percentage of nodes in the quadtree, which decorated with the model, is low and size of these nodes are small. So, the model has minimum influence in encoding rate. Also, because the \psml\ model takes more bits than the wedgelet model, images encoded with Wedgelets Transform has smaller size than the images encoded by the \psml\ algorithm in very low compression rates. This corollary is visible in right side of Fig. \ref{fig:WMrdPIC1bpp}.
By increasing the compression rate, the percentage of nodes in the quadtree which decorated with the model is increased and size of these nodes become larger. Because the \psml\ model is more fitted to fingerprint structure than wedgelet model, in not so small node, the rate-distortion ratio made by this model has higher values. So, in this range of rates, the \psml\ algorithm has advantage Wedgelets Transform in PSNR value. Also, by increasing the compression rate, this advantage become more pronounced. This corollary is visible in Fig. \ref{fig:WMrd}.
The noise and complexity in fingerprint images increases the percentage of nodes in quadtree decorated with the model. As the images of DB2 and DB3 have such a situation, in encoded images of these databases, the PSNR value results by \psml\ is higher than the PSNR value resulted by Wedgelets Transform.
In Fig. \ref{fig:WMpic}, some example of encoded images at two different compression rates are shown. The encoded images by \psml\ algorithm has better visual quality and minutiae is better identifiable, specially at higher compression rates.

\begin{figure*}[!t]
\centering
\subfloat[]{\includegraphics[width=0.9\columnwidthe,trim=0.8cm 0.2cm 1cm 0cm,clip]{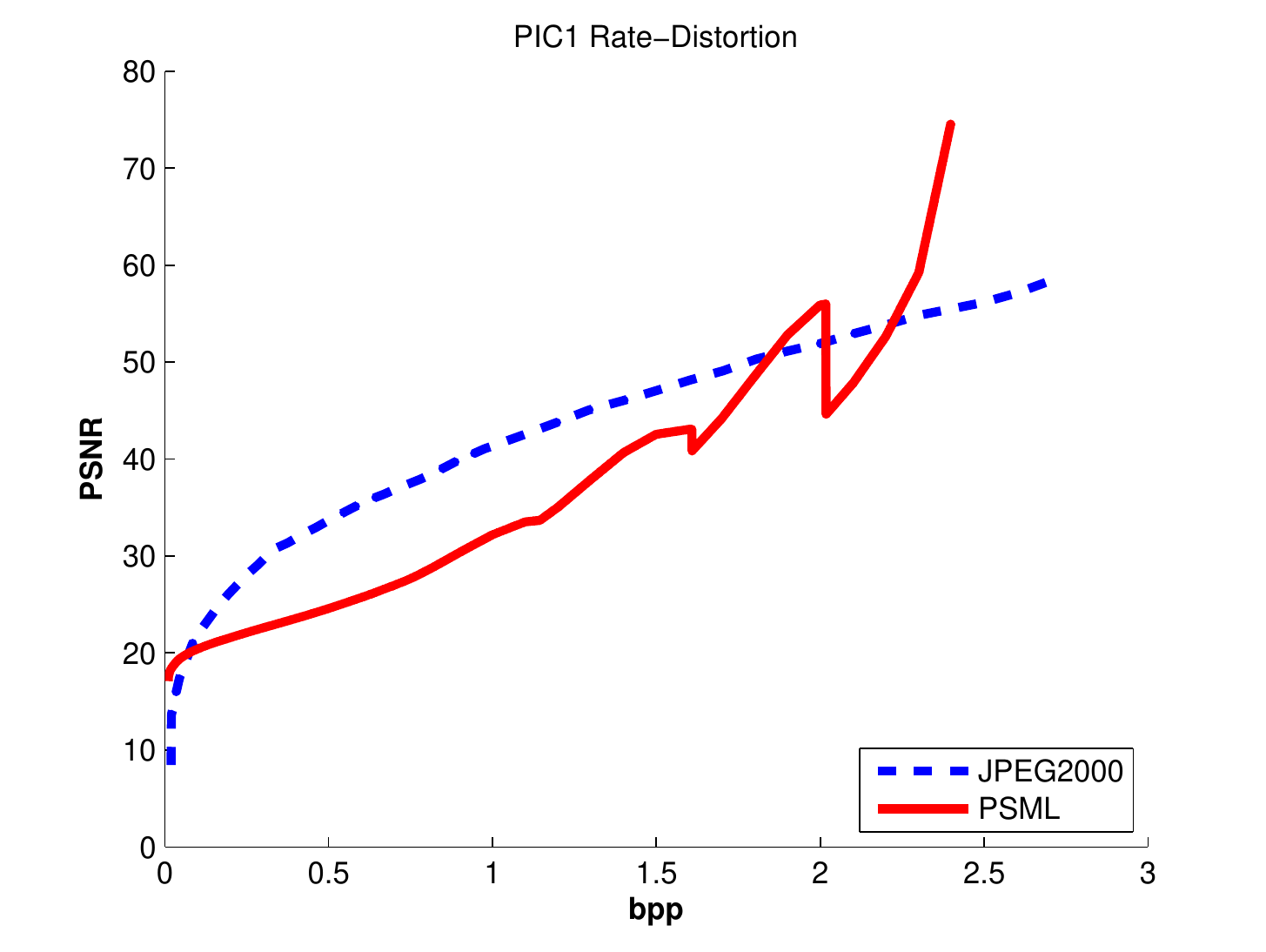}}
\hfill
\subfloat[]{\includegraphics[width=0.9\columnwidthe,trim=0.8cm 0.2cm 1cm 0cm,clip]{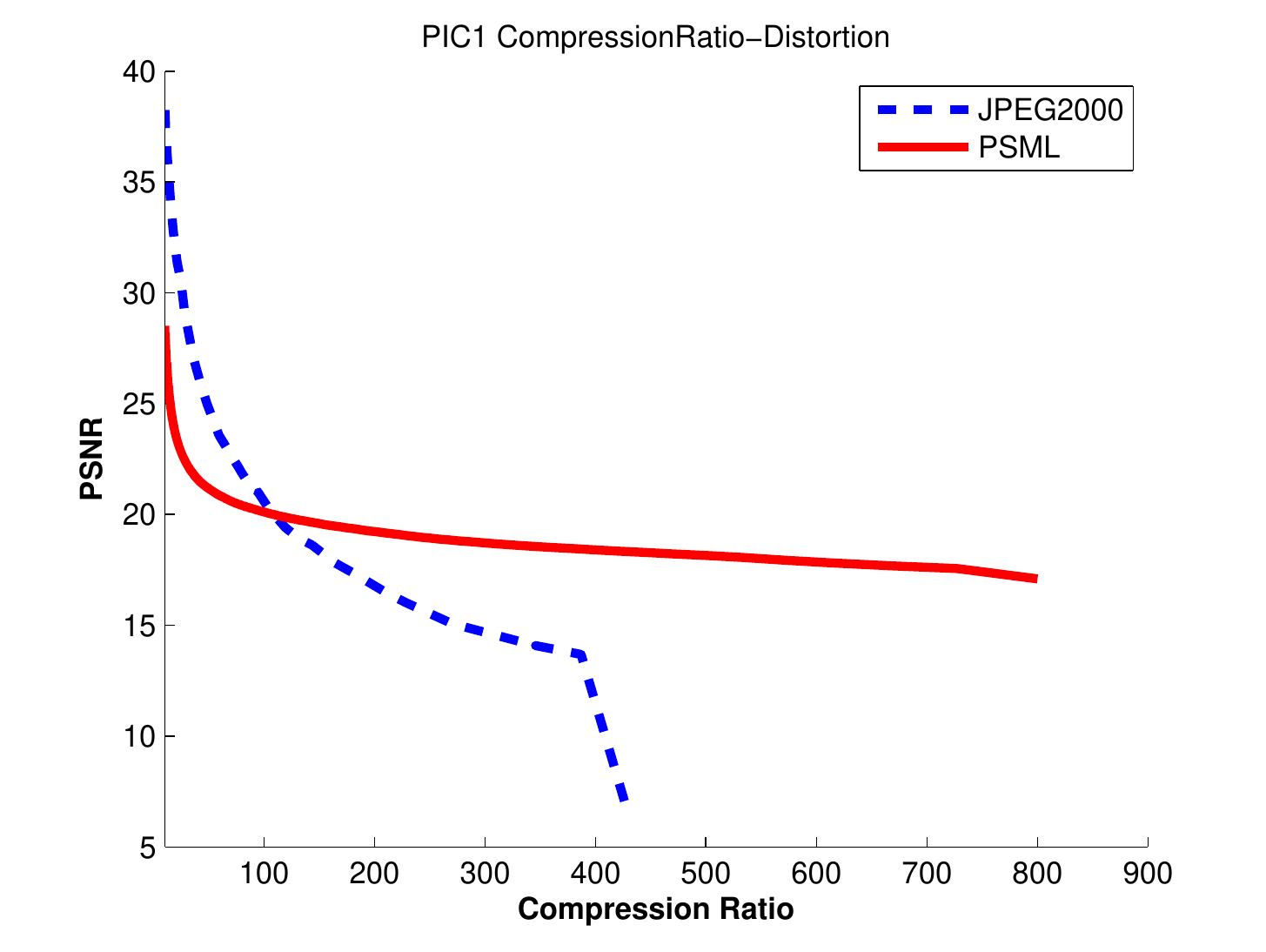}}
\\
\subfloat[]{\includegraphics[width=0.9\columnwidthe,trim=0.8cm 0.2cm 1cm 0cm,clip]{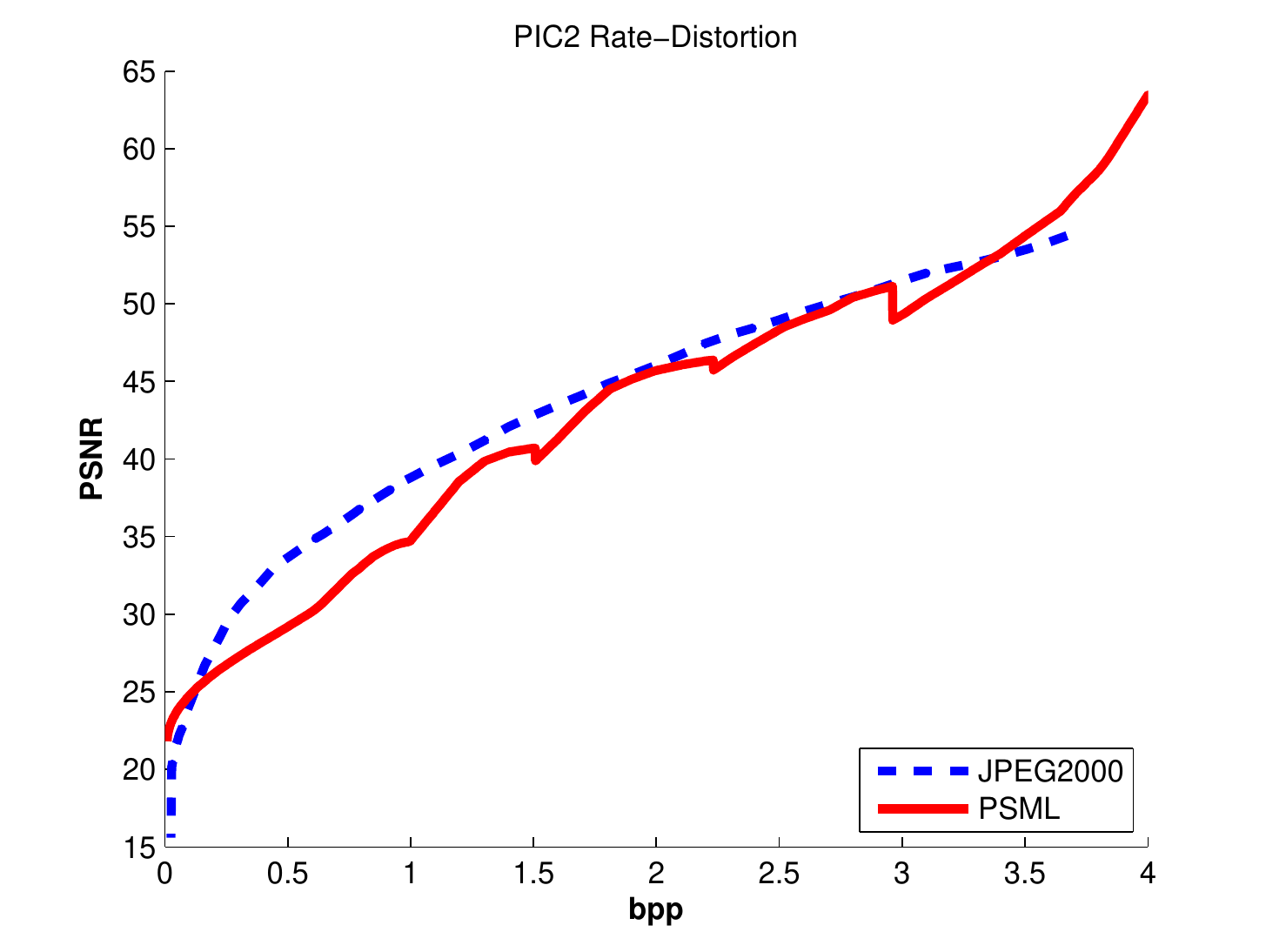}}
\hfill
\subfloat[]{\includegraphics[width=0.9\columnwidthe,trim=0.8cm 0.2cm 1cm 0cm,clip]{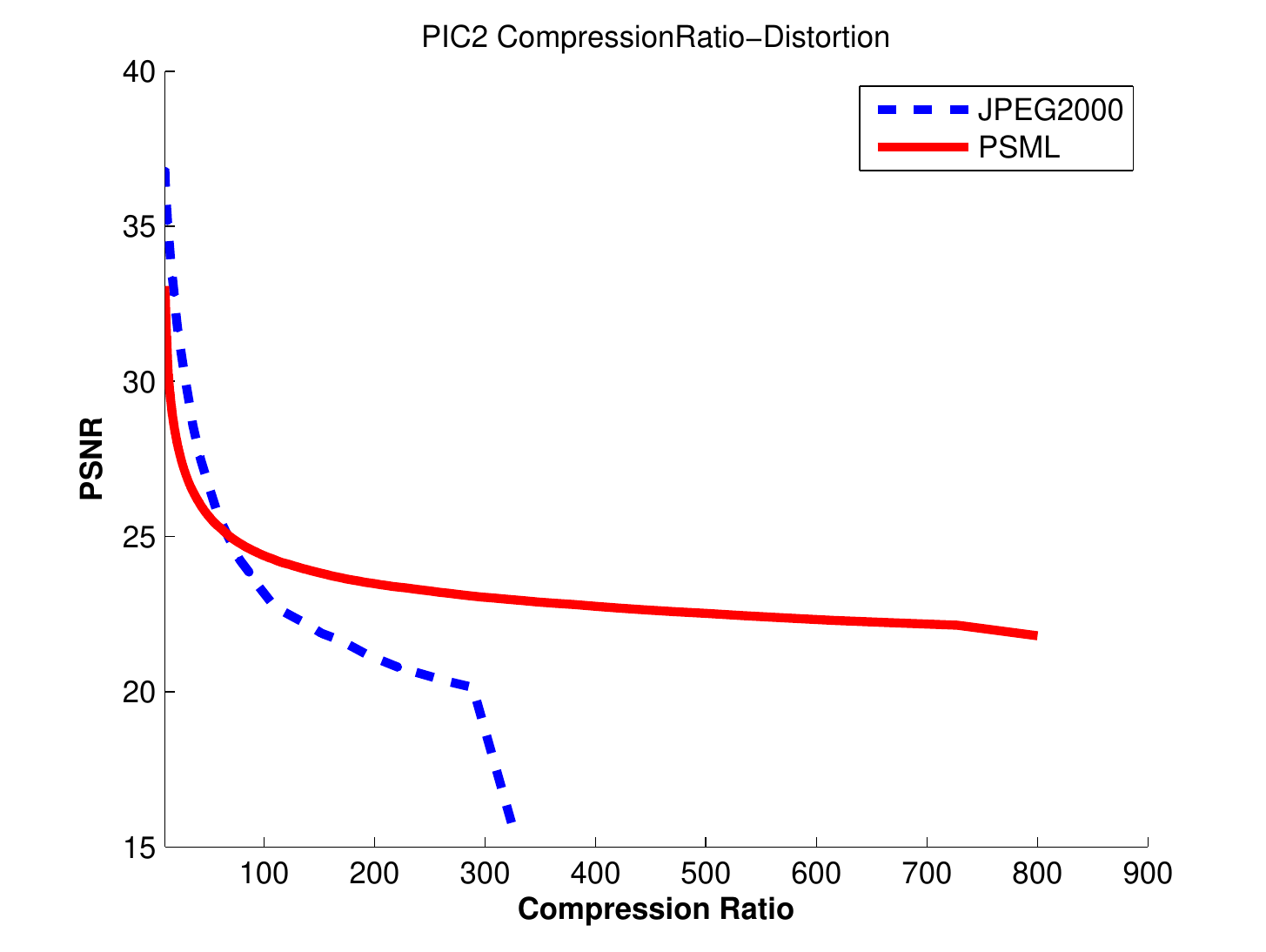}}
\\
\subfloat[]{\includegraphics[width=0.9\columnwidthe,trim=0.8cm 0.2cm 1cm 0cm,clip]{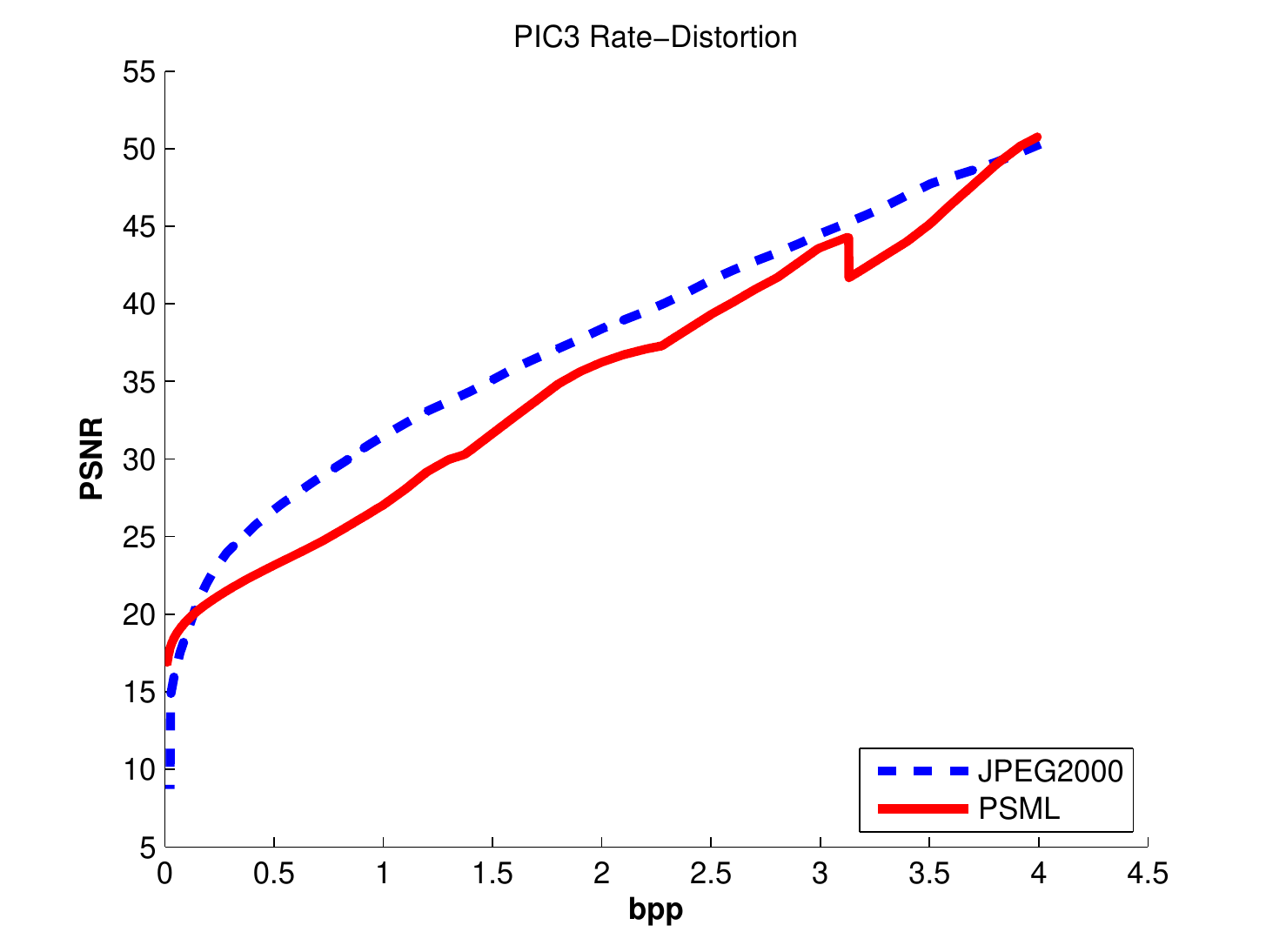}}
\hfill
\subfloat[]{\includegraphics[width=0.9\columnwidthe,trim=0.8cm 0.2cm 1cm 0cm,clip]{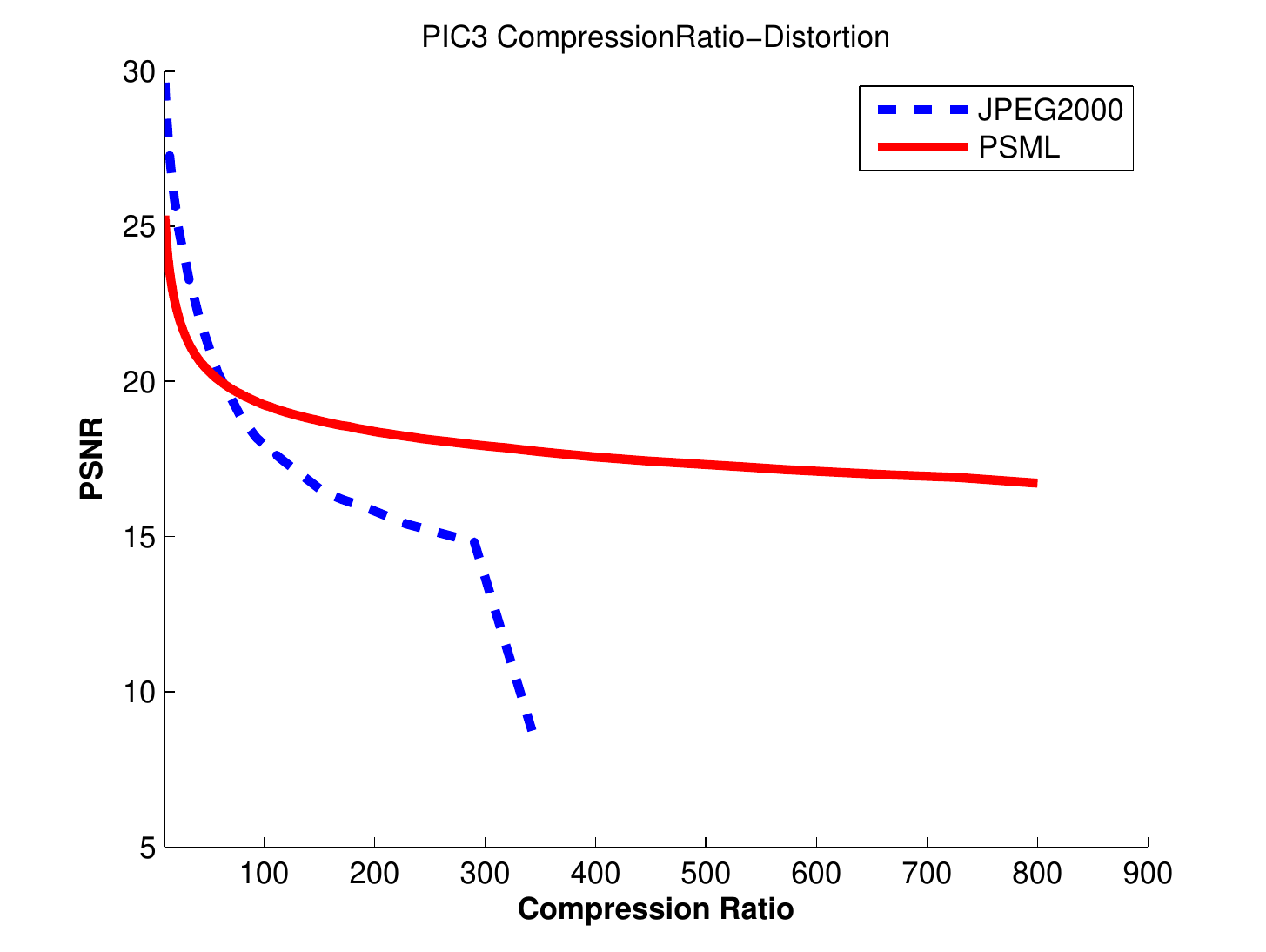}}
\caption{Compare rate-distortion (left) and compression ratio-distortion (right) for JPEG2000 algorithm and \psml\ algorithm on PIC3 according to experiment 2. Rate-distortion diagram shows results of low compression rates in detail, while compression ratio-distortion diagram shows result of high compression rates in detail. (a,b) PIC1 (c,d) PIC2 (e,f) PIC3.}
\label{fig:JMrd}
\end{figure*}
\begin{figure*}[!t]
\centering
\makebox[0.3\columnwidthe][c]{\tiny{Original image}} \quad \makebox[0.3\columnwidthe][c]{\tiny{JPEG2000 (1 bpp)}} \quad \makebox[0.3\columnwidthe][c]{\tiny PSML (1 bpp)} \quad \makebox[0.3\columnwidthe][c]{\tiny JPEG2000 (0.1 bpp)} \quad \makebox[0.3\columnwidthe][c]{\tiny PSML (0.1 bpp)}
\\
\subfloat[]{\includegraphics[width=0.3\columnwidthe]{DB1-101_1.png}}
\quad
\subfloat[]{\includegraphics[width=0.3\columnwidthe]{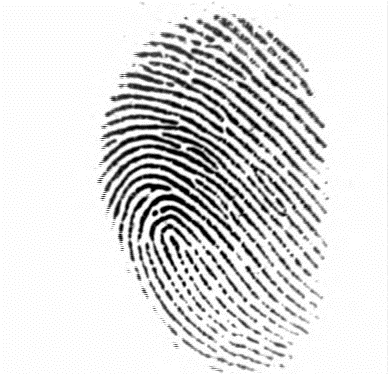}}
\quad
\subfloat[]{\includegraphics[width=0.3\columnwidthe]{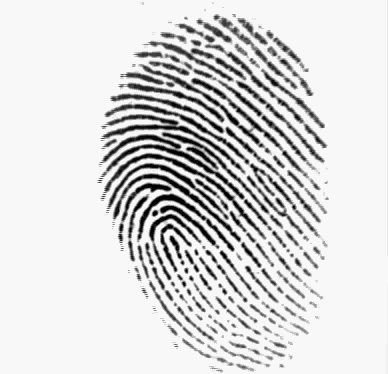}}
\quad
\subfloat[]{\includegraphics[width=0.3\columnwidthe]{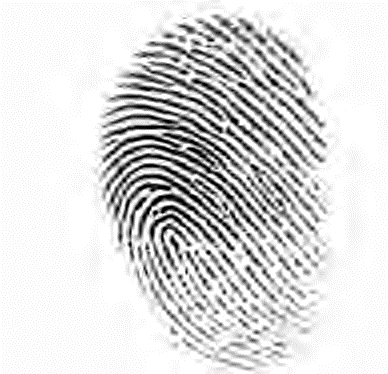}}
\quad
\subfloat[]{\includegraphics[width=0.3\columnwidthe]{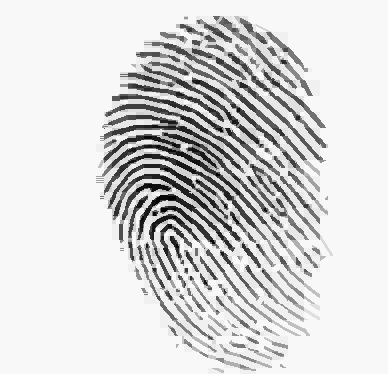}}
\\
\subfloat[]{\includegraphics[width=0.3\columnwidthe]{DB4-102_1.png}}
\quad
\subfloat[]{\includegraphics[width=0.3\columnwidthe]{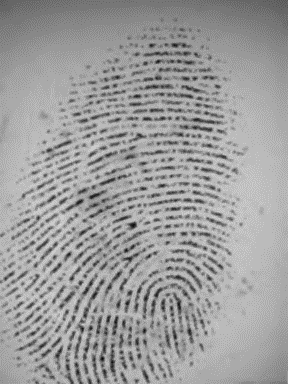}}
\quad
\subfloat[]{\includegraphics[width=0.3\columnwidthe]{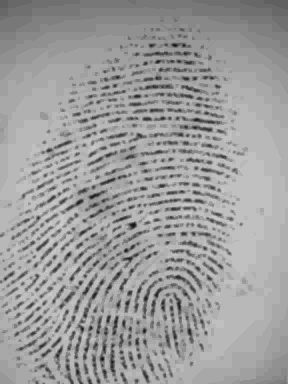}}
\quad
\subfloat[]{\includegraphics[width=0.3\columnwidthe]{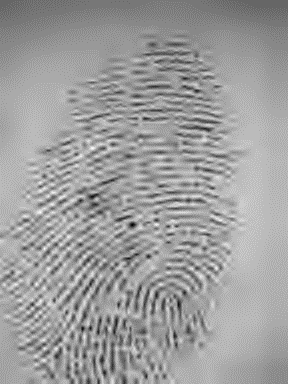}}
\quad
\subfloat[]{\includegraphics[width=0.3\columnwidthe]{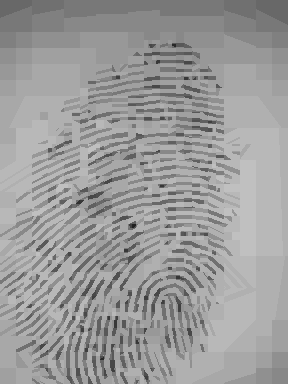}}
\\
\subfloat[]{\includegraphics[width=0.3\columnwidthe]{DB5-012_1_2.png}}
\quad
\subfloat[]{\includegraphics[width=0.3\columnwidthe]{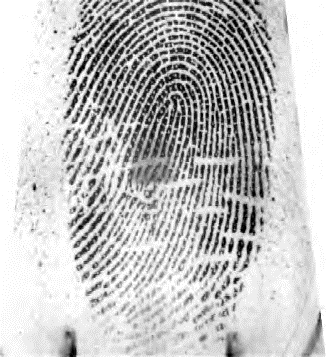}}
\quad
\subfloat[]{\includegraphics[width=0.3\columnwidthe]{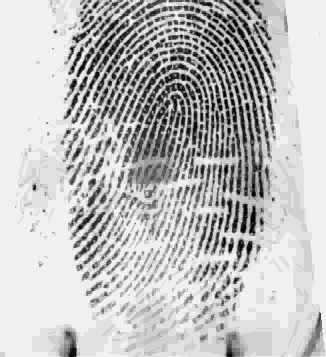}}
\quad
\subfloat[]{\includegraphics[width=0.3\columnwidthe]{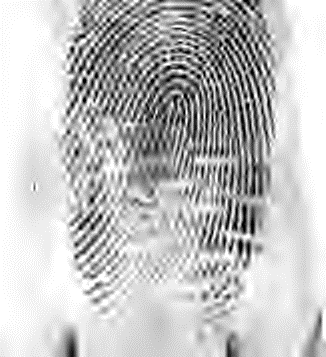}}
\quad
\subfloat[]{\includegraphics[width=0.3\columnwidthe]{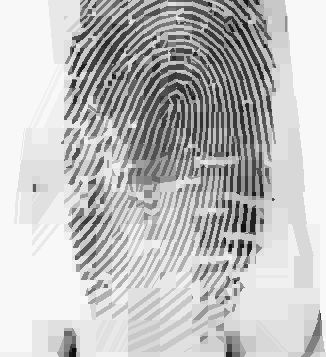}}
\caption{Samples of fingerprint images encoded by JPEG2000 and \psml\ algorithm, according to Experiment 2. First column shows the original fingerprint image. Second and fourth columns show the compressed fingerprint images by JPEG2000 algorithm. Third and Fifth columns show the compressed fingerprint images by \psml\ algorithm. Second and Third columns are encoded in 1 bpp, while fourth and fifth column are encoded in 0.1 bpp. First row belongs to result of experiment 1 on PIC1, while second and third rows belongs to PIC2 and PIC3, respectively. The PSNR value of images are (b,c) 41.36 , 32.19 (d,e) 21.90 , 20.41 (g,h) 38.79 , 34.74 (i,j) 24.22 , 24.75 (l,m) 31.49 , 27.03 (n,o) 18.73 , 19.58.}
\label{fig:JMpic}
\end{figure*}

\textbf{Experiment 2}: All images in DB1, DB2 and DB3 are encoded with JPEG2000 and \psml\ algorithm at different compression rates ranging from 0.01 to 4 bpp. The \psml\ algorithm builds quadtree up to 7 levels for model. Also, it uses arbitrary bit rates from 3 to 8 bits for gray level values (i.e., the same bit rate for all value). Also, we apply the compression algorithm proposed in section \ref{sec:modelcompress}. According to results, almost in all images, the same result was obtained. Similar to experiment 1, the rate-distortion and compression ratio-distortion diagrams for PIC1, PIC2 and PIC3 are shown in Fig. \ref{fig:JMrd}. In rate-distortion diagrams of this figure, PSML algorithm has some jumps, because the entire graylevel values in image is quantized with the same number of bits. So, when number of bits changed from 8 to 7, 7 to 6 or etc., it is observed as a jump in the PSNR value. Also, Fig. \ref{fig:JMpic} shows the compressed images obtained by both algorithms at two different rates, 0.1 bpp and 1 bpp.

%
%

\begin{figure*}
\centering
\subfloat[]{\includegraphics[width=0.95\columnwidthe,trim=0.8cm 0.2cm 1cm 0cm,clip]{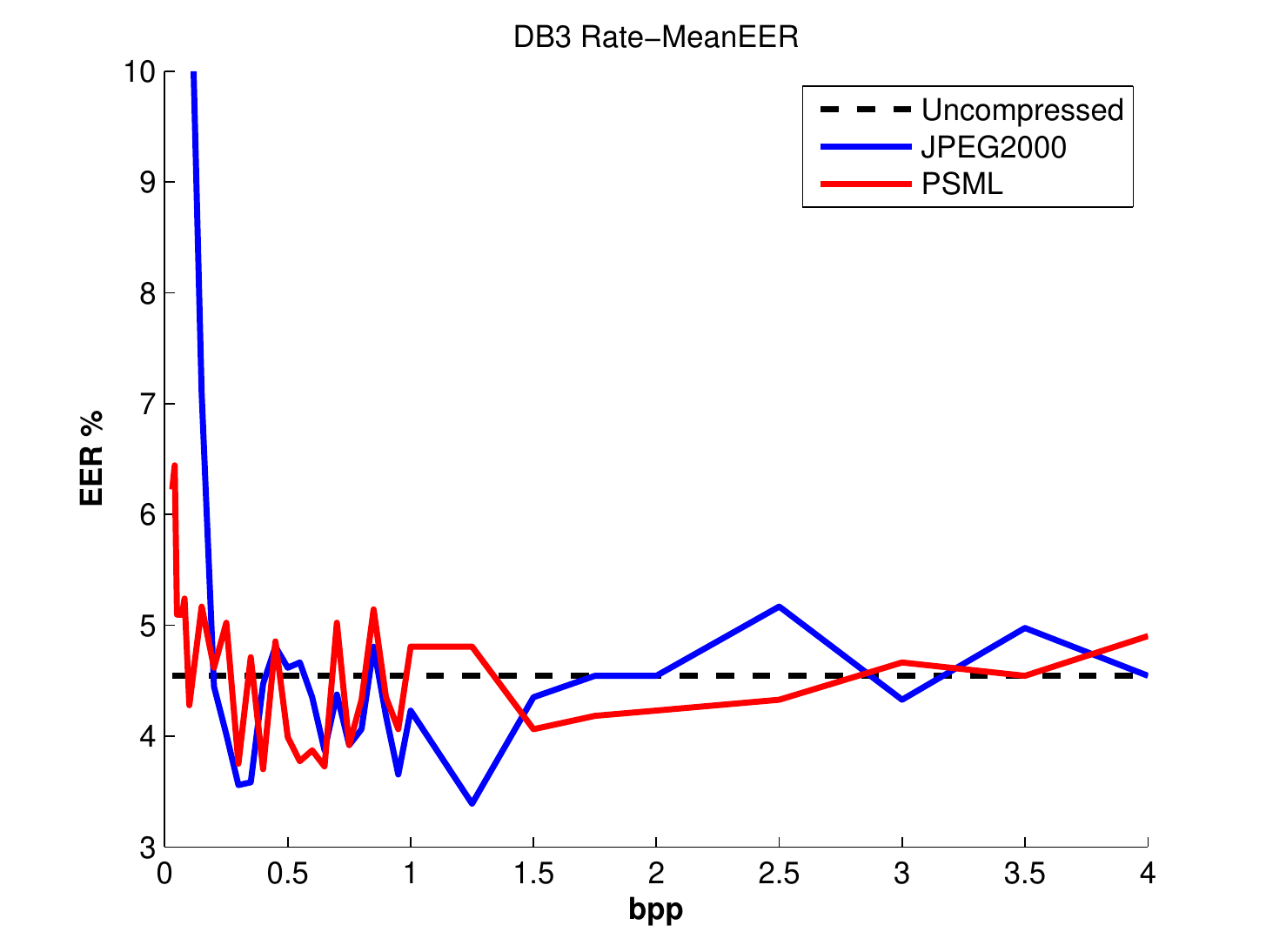}}
\hfill
\subfloat[]{\includegraphics[width=0.95\columnwidthe,trim=0.8cm 0.2cm 1cm 0cm,clip]{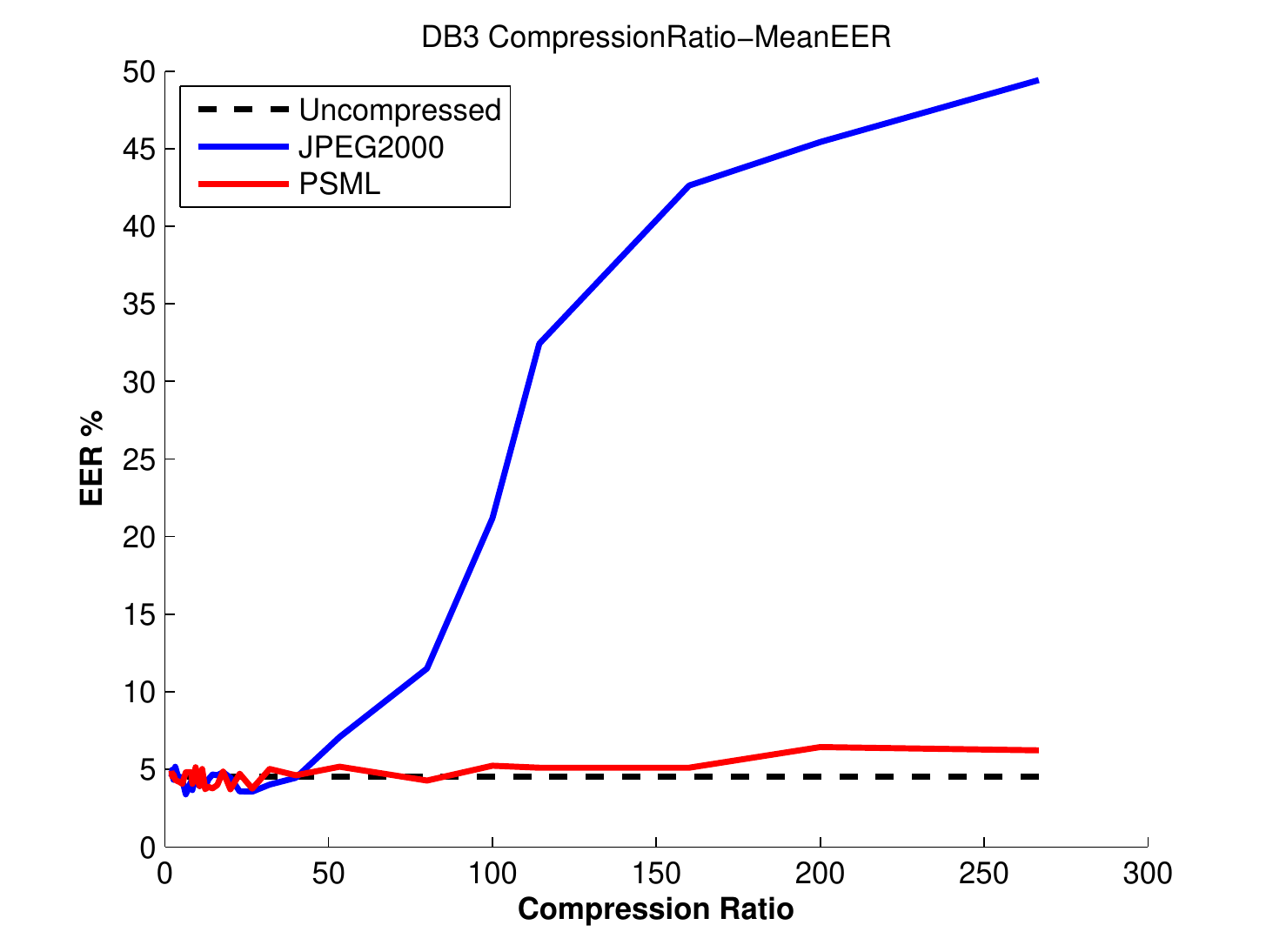}}
\caption{The EER values obtained by experiment 3 for various compression rates. The dashed line show the EER value of uncompressed images. The thiner and thicker lines show EER values of JPEG2000 and \psml\ algorithms, respectively.}
\label{fig:JMeer}
\end{figure*}

\textbf{Experiment 3}: In this experiment, the effect of compression algorithms examined with an AFIS tool. This experiment applied to DB3 images, examined \psml\ and JPEG2000 compression algorithms and repeated for various number of compression rates. At each compression rate and for each compression algorithm, all images are encoded in specified rate. Then the VeriFinger tool is applied to all compressed images to obtain matching value between each two pair of images. Finally, for each compression rate and for each compression algorithm, the FAR/FRR diagram was obtained.
From each diagram, the EER value can be extracted. Fig. \ref{fig:JMeer} shows the mean EER value in different compression rates for both compression algorithms.


\textbf{Corollary 2 (JPEG2000 vs \psml)}: The resulting diagrams of experiment 2 (Fig. \ref{fig:JMrd}), can be divided into three parts. In very low compression rates (compression ratio less than 3:1), the PSNR value of the \psml\ algorithm has advantage over JPEG2000 algorithm. This is because most nodes in quadtree are constant, so the coefficient compression algorithm acts like JPEG-LS. In middle range of compression rates (compression ratio greater than 3:1 and less than 80:1), the PSNR value of JPEG2000 algorithm has advantage over the \psml\ algorithm. But, as can be seen in Fig. \ref{fig:JMeer}, the mean EER value of JPEG2000 does not have this advantage. At compression rates less than 40:1, the mean EER value of both algorithms is approximately equal and at compression rates greater than 40:1, the \psml\ algorithm has advantage over JPEG2000 algorithm in terms of mean EER value. This implies that although the \psml\ algorithm does not preserve gray level value of each pixels, but it preserves the discriminating information required by the VeriFinger tool to compute the matching result between two fingerprint images. It is clear that the information is fingerprint structure and minutiae. In very high compression rate (compression ratio greater than 80:1), the PSNR value and mean EER value of \psml\ algorithm has great advantage over JPEG2000 algorithm. It implies that for higher compression rates, the \psml\ algorithm is a much better choice than JPEG2000 for compression fingerprint images.

\textbf{Corollary 3 (\psml\ robustness)}: By analyzing the results of experiment 3 (Fig. \ref{fig:JMeer}), we conclude that the \psml\ algorithm is much more robust in terms of preserving fingerprint information than JPEG2000 algorithm. In common range of compression ratio (less than 40:1) neither \psml\ algorithm nor JPEG2000 has advantage over another one. But in higher compression rates, mean EER value of JPEG2000 is close to 50\%, that means identification algorithm works like a random algorithm and the compressed images are not identifiable from each other. But, mean EER value of \psml\ algorithm remains in acceptable range and the compressed images identifiable from each other.

\begin{figure*}
\centering
\subfloat[]{\includegraphics[width=0.3\columnwidthe]{DB1-101_1.png}}
\quad
\subfloat[]{\includegraphics[width=0.3\columnwidthe]{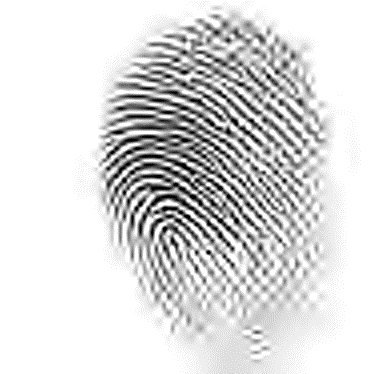}}
\quad
\subfloat[]{\includegraphics[width=0.3\columnwidthe]{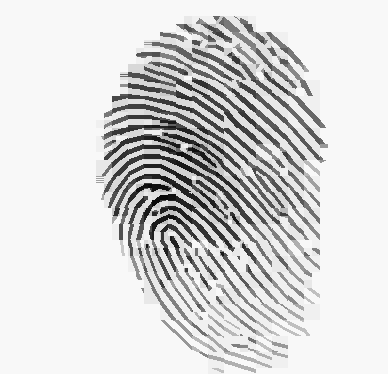}}
\quad
\subfloat[]{\includegraphics[width=0.3\columnwidthe]{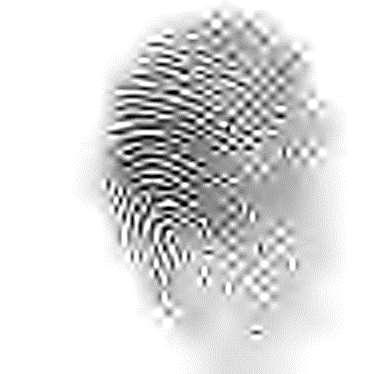}}
\quad
\subfloat[]{\includegraphics[width=0.3\columnwidthe]{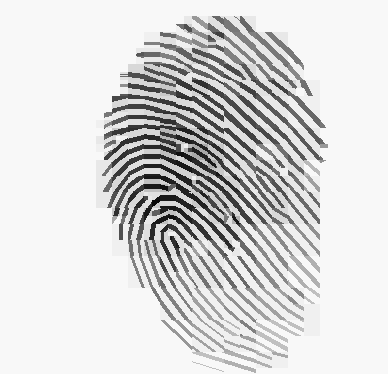}}
\caption{Samples of compressed fingerprint images in high compression rates. (a) PIC1 original image (b,d) encoded with JPEG2000 (c,e) encoded with \psml. Compression ratio is (b,c) 178:1 (d,e) 320:1.}
\label{fig:HighCR}
\end{figure*}

\textbf{Corollary 4 (Fingerprint image compression in high compression ratio)}: At high compression ratios, fingerprint images encoded by JPEG2000 lose their structure and discriminating information. But, if they are compressed by \psml\ algorithm, the fingerprint discriminating information is better retained. Fig. \ref{fig:JMrd} shows JPEG2000 is not have ability to encode fingerprint images in compression ratio greater than 200:1 and the PSNR value is fall sharply at these compression ratios. But, the quality of the compressed images by JPEG2000 decreases before this range of compression ratios. This corollary is confirmed by Fig. \ref{fig:HighCR}.

\section{Conclusion and future works}

A simple structure called \psml\ model is proposed. By using this model in compressing algorithm, the structure of fingerprint images are preserved even in very high compression rates. In comparison with the most known algorithm in its family, the experiments show that the \psml\ algorithm has better PSNR values in comparison with Wedgelets Transform.

One of the most important issues that must considered when developing compression algorithm for fingerprint images is preserving minutiae, which are used by AFIS systems for identification. In high compression rates, the JPEG2000 algorithm is ineffective, but the \psml\ algorithm can compress fingerprint images without losing structure and discriminating information of these images. According to the result of AFIS system applied on encoded images with \psml\ algorithm, the proposed method is preferred to JPEG2000 when high compression rates is needed.

There are several future directions that can follow this work. First, the compression algorithm that introduced in section \ref{sec:modelcompress} is very simple. It can be improved to achieve better compressing algorithm. As a result, the encoded images becomes more compressed while its advantages remains as before. Secondly, the proposed model could be extended to consider bifurcation and delta (Y-shaped ridge meeting) that are most important for minutiae extraction. Thirdly, other initialization methods can be investigated for better results. Fourthly, the fast approximation algorithm can be replaced either by an optimization algorithm that the solution is more closer to global optimal solution, or by a faster approximation algorithm. Finally, the proposed algorithm can be implemented in parallel to reduce the time complexity.

\section*{References}
\bibliography{bibliography}
\end{document}